\documentclass[10pt]{article} 

\usepackage[accepted]{tmlr}

\usepackage{amsmath,amsfonts,bm}

\def\eqref#1{equation~\ref{#1}}

\def\1{\bm{1}}

\DeclareMathAlphabet{\mathsfit}{\encodingdefault}{\sfdefault}{m}{sl}
\SetMathAlphabet{\mathsfit}{bold}{\encodingdefault}{\sfdefault}{bx}{n}

\usepackage[utf8]{inputenc} 
\usepackage[T1]{fontenc}    
\usepackage{hyperref}       
\usepackage{url}            
\usepackage{booktabs}       
\usepackage{nicefrac}       
\usepackage{microtype}      
\usepackage{xcolor}         
\usepackage{float}
\usepackage{newfloat}
\usepackage{placeins}
\usepackage[export]{adjustbox}
\usepackage{multirow}

\usepackage{arydshln}
\usepackage{amsmath}
\usepackage{amssymb}
\usepackage{amsfonts}       

\usepackage{graphicx}
\usepackage{wrapfig}
\usepackage{subcaption}

\title{Towards Efficient Mixture of Experts: A Holistic Study of Compression Techniques}

\author{\name Shwai He\thanks{Equal contribution} \email shwaihe@umd.edu \\
      \addr 
      University of Maryland, College Park
      \AND
      \name Daize Dong$^*$\thanks{Work done during assistantship} \email daize.dong@rutgers.edu \\
      \addr 
      Rutgers University
      \AND
      \name Liang Ding \email liangding.liam@gmail.com\\
      \addr University of Sydney 
      \AND
      \name Ang Li \email angliece@umd.edu\\
      \addr 
      University of Maryland, College Park \\
      }

\definecolor{color1}{rgb}{0.1,0.7,0.8} 
\definecolor{color2}{rgb}{0.9,0.1,0.1} 
\definecolor{color3}{rgb}{0.7,0.3,0.7} 
\definecolor{color4}{rgb}{0.3,0.3,0.7} 
\definecolor{color5}{RGB}{8, 102, 3} 
\definecolor{color6}{rgb}{0.53, 0.66, 0.42} 

\newcommand{\weight}{\boldsymbol{W}}
\newcommand{\expert}{\boldsymbol{E}}
\newcommand{\expertshared}{\bar{\boldsymbol{E}}}
\newcommand{\mask}{\boldsymbol{M}}
\newcommand{\x}{\boldsymbol{x}}
\newcommand{\y}{\boldsymbol{y}}
\newcommand{\X}{\mathcal{X}}
\newcommand{\T}{\mathcal{T}}

\newcommand{\score}{\boldsymbol{S}}
\newcommand{\gate}{\boldsymbol{G}}

\newcommand{\mixtral}{Mixtral-8$\times$7B}
\newcommand{\deepseek}{DeepSeek-MoE-16B}
\newcommand{\expertslimming}{Expert Slimming}
\newcommand{\experttrimming}{Expert Trimming}
\newcommand{\expertdrop}{Expert Drop} 
\newcommand{\layerdrop}{Layer Drop} 
\newcommand{\blockdrop}{Block Drop} 

\usepackage{pifont}
\newcommand{\cmark}{\ding{51}}
\newcommand{\xmark}{\ding{55}}
\newcommand{\omark}{\ding{109}}

\def\month{03}  
\def\year{2025} 
\def\openreview{\url{https://openreview.net/forum?id=HTpMOl6xSI}} 

\begin{document}

\maketitle

\begin{abstract}
Scaling large language models has driven remarkable advancements across various domains, yet the continual increase in model size presents significant challenges for real-world deployment. The Mixture of Experts (MoE) architecture offers a promising solution by dynamically selecting and activating only a subset of experts during inference, thus substantially reducing computational costs while preserving high performance. Despite these benefits, MoE introduces new inefficiencies, such as excessive parameters and communication overhead.
In this work, we present a holistic study of compression techniques for Mixture of Experts to enhance both efficiency and scalability. 
While recent efforts have focused on \experttrimming, which reduces the number of experts, these approaches still suffer from considerable communication and computational costs. 
To address this, we propose more aggressive strategies, such as \layerdrop, which removes entire MoE layers, and \blockdrop, which eliminates transformer blocks. Surprisingly, these aggressive pruning techniques not only preserve model performance but also substantially improve computation and memory efficiency.
Furthermore, beyond \experttrimming, we also introduce \expertslimming, which compresses individual experts to further boost performance and can be seamlessly integrated with \experttrimming. 
Extensive experimental results demonstrate the effectiveness of our proposed methods — \layerdrop~and \blockdrop~— along with the comprehensive recipe that integrates \expertslimming~and \experttrimming, achieving a $6.05\times$ speedup with $77.1$\% reduced memory usage while maintaining over $92\%$ of performance on \mixtral.
Our code is released at \url{https://github.com/CASE-Lab-UMD/Unified-MoE-Compression}. 
\end{abstract}

\section{Introduction}

While scaling large language models has shown exceptional performance across various domains \citep{ramesh2021zeroshot, openai2024gpt4, geminiteam2024gemini}, the increasing model size poses significant challenges in real-world deployments~\citep{wanda, gptq} due to excessive computational demands and associated costs.
The Mixture of Experts (MoE) \citep{shazeer2017outrageously}, which selectively activates a subset of parameters during inference, offers a promising solution to reduce these computational burdens. Additionally, integrating MoE with Large Language Models (LLMs) has been shown to further enhance performance \citep{jiang2024mixtral, dai2024deepseekmoe}.

Despite these advancements, MoE models still suffer from significant redundancies that increase deployment costs. Standard MoE implementations replicate feed-forward layers across multiple experts, resulting in heavily parameterized models. For instance, \mixtral~\citep{jiang2024mixtral} contains 47B parameters, but only 13B parameters are activated per token, leading to substantial GPU memory consumption and limited scalability. In addition, replicating experts often introduces redundant experts. 
For example, \cite{he-etal-2023-pad} observed that expert parameters could be compressed through parameter sharing. Similarly, \cite{lu2024experts} noted that not all experts are essential, suggesting that some can be safely removed. These findings underscore the potential for compressing MoE models to improve efficiency without sacrificing effectiveness. 

In this paper, we first investigate the \experttrimming~based compression techniques, which reduce the number of experts to enhance the efficiency of MoE \citep{cheng2020survey, liang2021pruning}. 
The most prevalent approach for \experttrimming~is \expertdrop, which scores each expert and drops the less important ones \citep{lu2024experts, muzio2024seermoe}.  
While \expertdrop~reduces model size, it does not eliminate costly computations within the MoE layer and complex communication among experts, leading to negligible improvements in the inference speed. 
To this end, we propose more aggressive \experttrimming~methods to enhance MoE efficiency. Specifically, to mitigate communication and computation costs, we present \layerdrop~that removes the entire MoE layer. Additionally, given the computation-intensive nature of the attention mechanism within transformer blocks, we further propose \blockdrop, which removes the whole transformer blocks.
We use similarity-based metrics to demonstrate the feasibility of \layerdrop~and \blockdrop. Surprisingly, these two coarse-grained methods outperform fine-grained \expertdrop~by a large margin in balancing performance and efficiency.
Additionally, with small-scale post-finetuning, the compressed models can be further optimized to achieve near-original performance.

Beyond removing experts, we also explore \expertslimming, which focuses on compressing individual experts. Techniques such as network pruning \citep{han2016deep, zhu2017prune} and quantization \citep{jacob2017quantization, nagel2021white} have proven effective for the model compression, with quantization being particularly suitable for hardware acceleration. By integrating \expertslimming~with \experttrimming, we propose a unified framework for compressing MoE models that further maximizes efficiency gains while maintaining strong performance.

Our experimental results on representative MoE models, \mixtral~\citep{jiang2024mixtral} and \deepseek~\citep{dai2024deepseekmoe}, demonstrate the effectiveness of our proposed methods. 
For \experttrimming, \expertdrop~significantly reduces the memory usage but it provides only marginal improvements in inference speed. In contrast, \layerdrop~and \blockdrop~significantly accelerate inference and reduce memory usage while maintaining comparable performance to the original models. 
The combined strategy of \experttrimming~and \expertslimming~results in a 6.05$\times$ speedup with only 22.8\% memory usage (20.0GB) while maintaining over 92\% of the original performance on \mixtral. The findings offer valuable insights for enhancing the efficiency of MoE models. 
Additionally, post-finetuning allows compressed models to recover most of their original performance, resulting in a minimal 0.6\% performance gap compared to the uncompressed \deepseek~model.

In summary, by conducting a holistic study on compressing Mixture of Experts, our key contributions are:  

\begin{itemize}

\item 
We extend \experttrimming~to a higher architectural with \layerdrop~and \blockdrop, significantly enhancing computation and memory efficiency while preserving model performance. 

\item 
We integrate \experttrimming~with \expertslimming~to further achieve efficiency gains without compromising performance. 

\item 
Extensive experimental results demonstrate the effectiveness of our proposed methods, achieving a $6.05\times$ speedup and reducing memory usage to just $20.0$ GB, all while maintaining over $92\%$ of performance on \mixtral.

\end{itemize}

\section{Related Work}


\paragraph{Mixture of Experts}
The Mixture of Experts (MoE) is a kind of neural network architecture with an extended set of parameters (referred to as ``experts'') controlled by a router, which is first introduced in the context of conditional computation \citep{jacobs1991adaptive, jordan1994hierarchical}.
The potential of sparse activation in MoE is subsequently exploited by \cite{shazeer2017outrageously} for efficient training and inference on pretrained models with special designs, opening the door for MoE in various vision \citep{riquelme2021scaling} and language \citep{lepikhin2020gshard, du2022glam, fedus2022switch} scenarios.
Attributed to its exceptional efficiency, MoE has been adopted as a foundational framework in the designs of large language models (LLMs) \citep{jiang2024mixtral, dai2024deepseekmoe, xue2024openmoe, llama-moe-2023, qwen_moe}, achieving superior scaling laws at low computational costs \citep{clark2022unified}.
Further investigations emerge in developing improved expert structures \citep{gururangan2022demix, rajbhandari2022deepspeed, dai2024deepseekmoe}, router designs \citep{lewis2021base, roller2021hash, zhou2022mixture}, and training strategies \citep{shen2023moduleformer, chen2022sparse}, propelling the continuous evolution on the representation capability and computational efficiency of MoE models. 
Despite the success, MoE also suffers from efficiency issues. For instance, MoE replicates the experts, significantly increasing the parameter budget \citep{he-etal-2023-pad}. On the other hand, adopting multiple experts to process input tokens introduces communication costs and enhances latency \citep{song2023powerinfer, xue2024moeinfinity}.

\paragraph{Compression Methods}
The increasing size of large language models presents considerable challenges for their practical implementation. Consequently, a range of efficient methods has emerged to address the implementation issues. Among them, model quantization \citep{gptq, awq} and network pruning \citep{wanda, sparsegpt} are widely utilized. Model quantization reduces the precision of neural network weights to lower bits \citep{jacob2017quantization}, while network pruning \citep{han2016deep} removes redundant parameters or architectures. Although these methods have shown promising results on dense models, they lack consideration for the inductive bias inherent in MoE. 
To bridge this gap, \expertdrop, as proposed in studies like \citep{muzio2024seermoe, lu2024experts}, addresses the unique nature of MoE by removing unimportant experts. By eliminating redundant experts, the MoE architecture becomes more compact and can be deployed at a lower cost. However, while \expertdrop~leads to a more compact architecture, it may also lead to non-negligible performance drop and rely on post-training for recovery.

\section{Preliminaries}





\subsection{Mixture of Experts}
A Mixture of Experts (MoE) layer consists of a collection of $n$ experts, $\{\expert_1, \expert_2, \dots, \expert_n\}$, each associated with weights $\{\weight_1, \weight_2, \dots, \weight_n\}$, and a router $\gate$ that dynamically selects the most relevant experts for a given input $\x$. The router computes selection scores, $\gate(\x) \in \mathbb{R}^n$, for all experts and selects the top $k$ experts, resulting in a sparse activation pattern. The input $\x$ is processed by the selected experts, and their outputs are combined into a weighted sum based on the router’s scores. This process is mathematically expressed as:
\begin{equation}
    \mathcal{K} = \mathrm{TopK}(\mathrm{Softmax}(\gate(\x)), k),
\end{equation}
\begin{equation}
    \y = \sum\nolimits_{i\in\mathcal{K}} {\gate(\x)_i \cdot \expert_i(\x|\weight_i)},
\label{eq:moe}
\end{equation}
where $\mathcal{K}$ denotes the indices of selected experts, $\gate(\x)_i$ represents the selection score for the $i$-th expert, and $\expert_i(\x)$ is the output from the $i$-th expert.
In transformer models, the MoE layer usually replaces the feed-forward network (FFN). In this context, each expert functions as an independent FFN module, enhancing the model's capacity without a proportional increase in the computational cost \citep{vaswani2017attention}.

\paragraph{Challenges} While MoE models have demonstrated strong performance across various tasks \citep{jiang2024mixtral, dai2024deepseekmoe}, they also encounter significant deployment challenges. On one hand, MoE models replicate multiple expert networks, inflating the model size and memory usage. For instance, \mixtral~has a total of 47B parameters and requires 87.7GB of memory for deployment, though only 13B parameters are activated for each token. On the other hand, the communication required to manage multiple expert networks increases latency and slows down inference speed, especially in distributed environments \citep{song2023powerinfer, yu2024moesysdistributedefficientmixtureofexperts}. 



\subsection{Overview of Previous Compression Methods}\label{subsec: overview}

To address the efficiency challenges, we first review several mainstream and state-of-the-art compression techniques for MoE models.

\textbf{Pruning}: Pruning reduces the number of active parameters by selectively disabling parts of the model’s weights. In an MoE layer with $n$ experts $\{\expert_i\}_{i=1}^{n}$ and corresponding weights $\{\weight_i\}_{i=1}^{n}$, pruning introduces binary masks $\{\mask_i\}_{i=1}^{n}$ to deactivate certain weights:
\begin{equation}
\label{eq:pruning}
    \hat \weight_i = \mask_i \odot \weight_i. 
\end{equation}
Pruning can be \textit{unstructured} \citep{lee2021layeradaptive, bai2022dual}, \textit{semi-structured}, or \textit{structured}. Unstructured sparsity tends to yield the best performance, semi-structured sparsity strikes a balance between efficiency and performance, and structured sparsity, while hardware-friendly, often results in lower performance. 

\textbf{Quantization}: Unlike pruning, which involves masking out unimportant parameters, quantization reduces memory usage by converting model weights to lower-bit representations. For MoE layers, quantization is applied in the following way:
\begin{equation}
\label{eq:quantization}
    \hat \weight_i = \text{Quant} ( \weight_i),       
\end{equation}
where ``$\mathrm{Quant}$'' denotes the quantization function. 
Quantization decreases the computation and memory consumption without reducing FLOPs or the total number of parameters, making it particularly advantageous for hardware acceleration. 

\textbf{\expertdrop}: Different from fine-grained pruning and quantization, \expertdrop~entails the removal of expert networks, based on the observation that not all experts are equally important \citep{lu2024experts, muzio2024seermoe}. Given expert-wise importance scores $\score$ 
(e.g., the routing scores, $\score(\expert_i) = \gate(\x)_i$), 
\expertdrop~retains only the experts with the highest $n'$ scores: 
\begin{equation}
\T' = \mathrm{TopK}(\score(\{\expert_i\}_{i=1}^{n}), n'), 
\end{equation}
\begin{equation}
\label{eq:expert_dropping}
    \expert \leftarrow \{\expert_i\}_{i \in \T'},    \quad\gate \leftarrow \gate_{i \in \T'}.
\end{equation}
Here, $\T'$ denotes the subset of the original expert indices $\T = \{1, 2, \dots, n\}$. \expertdrop~ reduces FLOPs conditionally: when $\T'$ contains more than or equal to $k$ indices, MoE still utilizes the top $k$ experts for each input; otherwise, it uses all remaining experts. While this approach reduces communication between experts, the resulting speedup is usually insignificant when maintaining acceptable performance.


\textbf{Other Compression Techniques}: 
Other methods, such as low-rank decomposition \citep{li2024merge, li2024lorap}, aim to compress model weights into smaller matrices, further reducing memory and computational costs. In this work, we primarily focus on the widely-used methods (pruning, quantization, and \expertdrop), leaving a more detailed exploration of these additional methods for future research.

\section{A Holistic Study of MoE Compression Techiniques}

\begin{figure}[]
    \centering    
    \includegraphics[width=\linewidth]{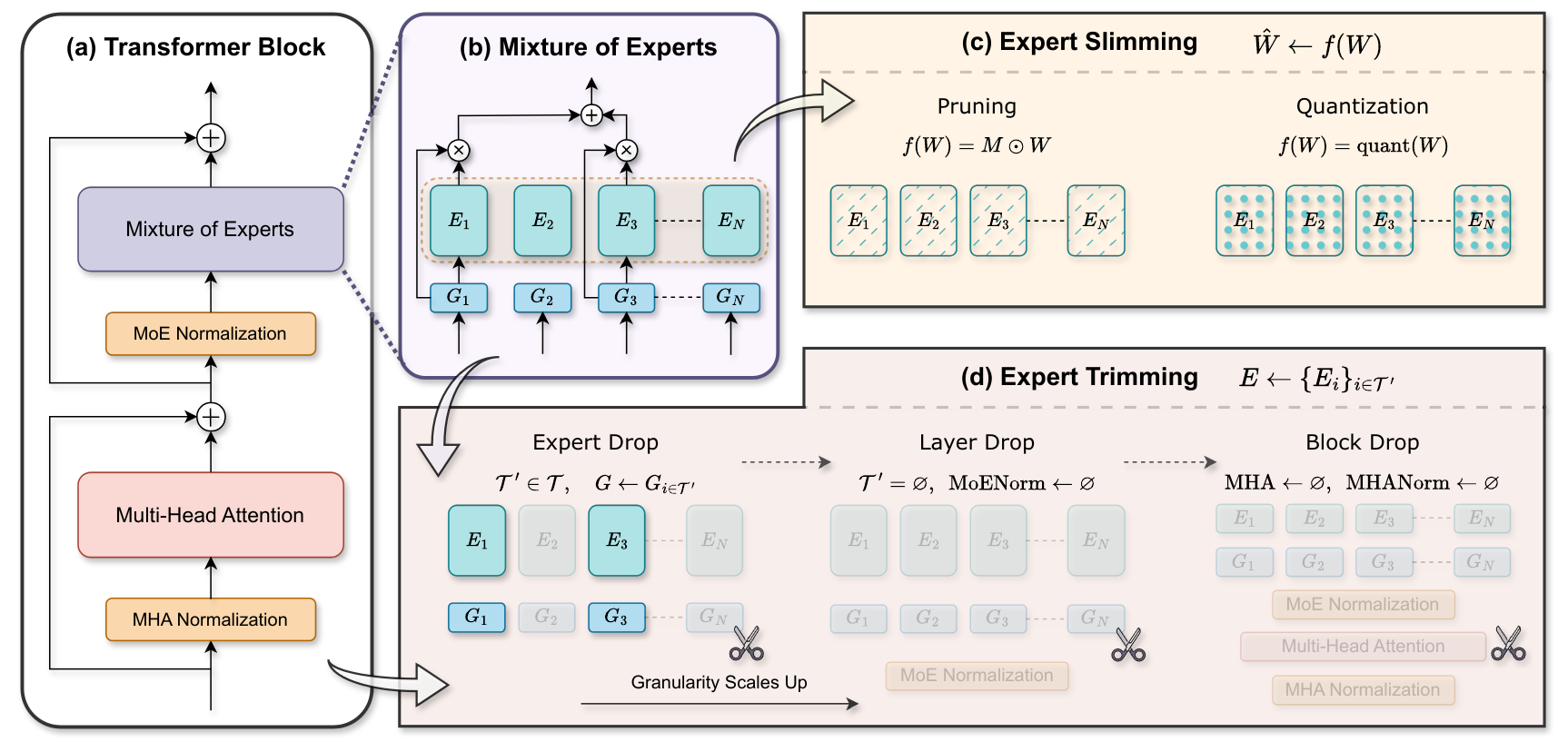}
\vspace{-3pt}
\caption{\textbf{The Unified View of MoE Compression.} The view integrates two complementary perspectives: \expertslimming~and \experttrimming. \expertslimming~compresses individual experts, while \experttrimming~directly drops structured modules.}
\label{fig:overview}
\end{figure}


In this section, we propose a general framework that unifies various compression methods for MoE. This framework provides a comprehensive understanding of MoE model efficiency issues and helps identify new design spaces for further performance improvements.

\subsection{Overview}
Existing MoE compression methods primarily address two types of inefficiencies: \textbf{structural redundancies} in the overall architecture and \textbf{internal redundancies} within individual experts. To address both issues, we categorize these methods into two complementary perspectives: \experttrimming~that focuses on removing structured components (e.g., experts, layers, and blocks), and \expertslimming~that compresses individual experts through techniques like pruning or quantization. 
An overview of these compression perspectives is illustrated in Figure \ref{fig:overview}.


\experttrimming~ deals with compressing structured modules by selecting and retaining only a subset of the experts, denoted as $\T’$. This is represented by the transformation $\T \leftarrow \T’$. Methods like \expertdrop, which selectively drops unimportant experts, are examples of this approach. 
On the other hand, the compression of individual experts (\expertslimming) focuses on the transformation and reduction of expert weights, denoted as $\weight$. We utilize a transformation function $f(\weight)$ to represent this process. The transformation function $f(\weight)$ can be understood as a general mapping that applies various compression techniques to the weights of the model. For example, in pruning, $f(\weight)$ could be a function that sets a subset of the weights to zeros. In quantization, $f(\weight)$ might reduce the precision of the weights from 32-bit floats to 8-bit integers. 
By integrating these two perspectives, we can derive a general form for efficient MoE models. The compression \textbf{within} and \textbf{across} experts can be expressed as follows:
\begin{equation}
    \y = \sum\nolimits_{i \in \T'} \gate_i \cdot \expert_i(\x | f(\weight_i)). 
    \label{eq:framework}
\end{equation}
In the following sections, we will elaborate on \experttrimming~and \expertslimming, respectively.

\begin{table}[]
\caption{
\textbf{Summary of Compression Methods.} ``\cmark''~means effective, indicating that the method performs well as intended. ``\xmark''~means ineffective, where the method fails to optimize the specified metric. ``\omark''~represents conditionally effective, depending on specific settings and environments.  
}
\vspace{-8pt}
\begin{center}
\resizebox{0.95\textwidth}
{!}{
        \begin{tabular}{lllcccc}
            \toprule
             & Method & Formulation & Parameter & Memory & FLOPs & Speedup \\ 
            \midrule

            & \expertdrop & $ \T \leftarrow \T'$ & 
            \cmark & \cmark & \omark & \omark \\ 
            \cdashline{2-7} 
            & \layerdrop & 
            & & & & \\ 
            \multirow{-3}{*}{\experttrimming} & \blockdrop & \multirow{-2}{*}{$\T \leftarrow \varnothing$} & 
            \multirow{-2}{*}{\cmark} & \multirow{-2}{*}{\cmark} & \multirow{-2}{*}{\cmark} & \multirow{-2}{*}{\cmark} \\
            \midrule 
            & Pruning & $\mask \odot \weight$ & 
            \cmark & \omark & \cmark & \omark \\
            \multirow{-2}{*}{\expertslimming} & Quantization & Quant($\weight$) & 
            \xmark & \cmark & \xmark & \cmark \\
            \bottomrule
        \end{tabular}
        }
    \end{center}
    \vspace{-10pt}
\label{tab:method_overview}
\end{table}

\subsection{\experttrimming}
The core operation of \experttrimming~involves updating the set of remaining experts denoted as $\T \leftarrow \T'$, where $\T'$ is a subset of the original expert indices $\T$. Specifically, \expertdrop~updates the experts and their corresponding routing weights as follows: $\expert \leftarrow \{\expert_i\}_{i \in \mathcal{T}'}$ and $\gate \leftarrow \gate_{i \in \T'}$.

However, \expertdrop~carries the risk of collapsing feature transformation. The absence of certain experts can lead to incorrect selections for given inputs, thereby degrading model performance \citep{chen2022sparse}. Additionally, partially reducing experts can disrupt routing patterns, negatively impacting the model's overall efficiency and effectiveness.
Despite its benefits, \expertdrop~still retains the costly computation within each expert and the complex communication between experts. These limitations highlight the need for further optimization of \experttrimming~to promote the efficiency. By systematically analyzing the redundancies and inefficiencies inherent in MoE models, 
we propose extending beyond expert-level optimizations to identify new design spaces for efficiency improvements.

\begin{wrapfigure}{rb}{0.42\textwidth}
    \vspace{-12pt}
    \centering 
\includegraphics[width=0.42\textwidth]{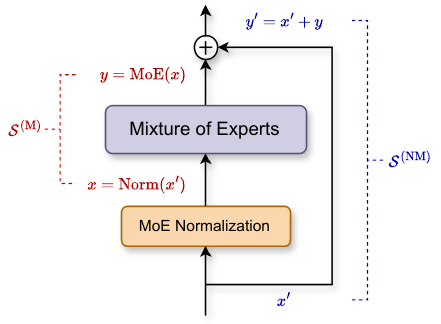} 
    \vspace{-18pt}
    \caption{\textbf{Illustration of Similarity Measurements in \layerdrop}. Features for calculating $\score^{\text{(M)}}$ and $\score^{\text{(NM)}}$ are colored with red and blue, respectively.}
    \vspace{-25pt}
\label{fig:layer_sim_detailed}
\end{wrapfigure}

We propose two novel techniques: \textbf{\layerdrop}~and \textbf{\blockdrop}. \layerdrop~focuses on removing entire MoE layers, which significantly reduces both computation and communication overhead. \blockdrop~extends this concept by eliminating entire blocks, including attention layers and MoE layers, within transformer models. These advanced techniques aim to streamline the model architecture, improve performance, and enhance overall efficiency.

\paragraph{\layerdrop}
Inspired by \cite{mod, layerskip}, we consider a special scenario of \expertdrop~where all experts are dropped ($\T \leftarrow \T' = \varnothing$), effectively removing entire MoE layers.
We refer to this approach as \layerdrop.  To perform \layerdrop, we use a similarity-based metric where high similarity indicates high redundancy in transformation.
One straightforward metric is the cosine similarity between the input $\x$ and the output $\y = \mathrm{MoE}(\x)$: 
\begin{equation}
    \begin{aligned}
        \score^{\text{(M)}} = \frac{\x \cdot \y}{||\x||_2 \ ||\y||_2}, ~~\text{where}~~\y = \mathrm{MoE}(\x).
    \end{aligned}
\label{eq:layerdrop_sim_moe}
\end{equation}
However, this metric alone does not adequately capture the impact of the MoE layer within the context of a transformer block, which includes a layer normalization module ("Norm") \citep{ba2016layer} and residual connections \citep{he2015deep}. 
To address this, we propose concurrently removing both the $\mathrm{MoE}$ and $\mathrm{Norm}$ layers. This approach ensures that the similarity metric more accurately reflects the combined functionality of these layers, allowing for a more precise identification of redundancy and a streamlined model architecture, as illustrated in Figure \ref{fig:layer_sim_detailed}. By considering the similarity between the raw residual input and the aggregated output, we can better evaluate the necessity of the MoE layer in the overall architecture:
\begin{equation}
    \begin{aligned}
 \score^{\text{(NM)}} = \frac{\x' \cdot \y'}{||\x'||_2 \ ||\y'||_2}, ~~       \text{where}~~\y'= \x' + \text{MoE}(\text{Norm}(\x')).
    \end{aligned}
\label{eq:layerdrop_sim_norm_moe}
\end{equation}
\paragraph{\blockdrop} 
Within a transformer block, \layerdrop~removes the MoE layers but retains the computation-costly attention layers \citep{ribar2024sparq, zhang2023h2o}. To address this issue, we further utilize the same similarity-based metrics to investigate whether the attention layer can be dropped without a significant performance drop. If feasible, this allows us to drop the entire block within MoE models, thus enhancing efficiency. 
We introduce \blockdrop~as an extension of \layerdrop, which also removes the attention layers. Specifically, for the $i$-th block, we assess its importance score by evaluating the similarity between its inputs $\x_l$ and outputs $\y_l$. 
Compared to \expertdrop, both \layerdrop~and \blockdrop~focus on structures beyond expert level, with the potential to further enhance the efficiency of MoE models. 


\subsection{\expertslimming}
Given that employing multiple experts in MoE significantly escalates parameters and inference costs, \expertslimming, stemming from single-model compression techniques, targets the compression of individual expert weights $\weight$ exclusively.
We denote any efficient transformation function as $f(\cdot)$, which encompasses pruning $\mask \odot \weight$ and quantization $\mathrm{Quant}(\weight)$.
Through the application of such functions, we reduce the redundancy within each expert and create several light-weighted slim experts, thus improving their intrinsic efficiency.
However, it is important to note that \expertslimming~primarily focuses on compressing individual experts without addressing the redundancy across multiple experts. For maximum efficiency gains, \expertslimming~and \experttrimming~can be integrated to compress both individual experts and structured components.
We summarize the efficiency contributions of all the discussed \experttrimming~and \expertslimming~methods in Table \ref{tab:method_overview}, highlighting the unique advantages of each approach. 

\section{Experiments on \experttrimming}

In this section, we evaluate the effectiveness of \experttrimming~techniques, starting with \expertdrop, and comparing it with our proposed methods, \layerdrop~and \blockdrop. 
Implementation details are provided in Appendix \ref{app:implementation_details}.


\label{sec:trimming}


\paragraph{\expertdrop: Performance Degradation with Limited Efficiency Gains}
While experts are specific structures in MoE, not all experts hold equal significance. Figure \ref{fig:expert_scores_distribution} visualizes the distribution of expert-wise importance scores, highlighting this variability. To systematically drop experts at varying proportions, we conduct experiments using both layer-wise and global dropping approaches (see Appendix \ref{app:expert_drop_details}). Given the importance of shared experts (Appendix \ref{sec:shared_experts}), we only dropped normal experts for \deepseek. Under both settings, \expertdrop~causes consistent performance degradation. For example, dropping $25\%$ of experts in \mixtral~results in a $23\%$ performance drop on the MMLU task.
The efficiency improvement from \expertdrop~is also marginal. For instance, dropping $12.5\%$ of experts results in less than a $1\%$ speedup, despite significant performance losses. More experimental results are available in Appendix \ref{sec:full_results}.

\begin{figure}[ht]
    \centering
    \includegraphics[width=0.94\linewidth]{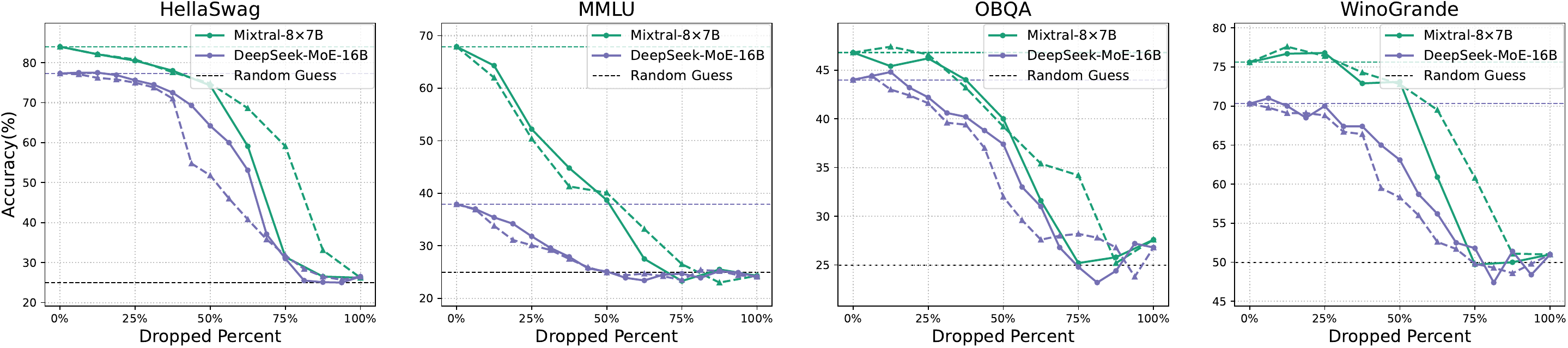} 
    \caption{\textbf{Evaluation of \expertdrop.} We consider two strategies: layer-wise dropping (dotted lines) and global dropping (solid lines). ``Random Guess'' refers to randomly generating an output rather than using the model’s predictions, serving as a baseline to assess the extent of performance degradation. 
    }
    \label{fig:expert_drop}
\end{figure}






\paragraph{\layerdrop: Comparable Performance with Greater Efficiency 
}

\begin{wrapfigure}{rb}{0.38\textwidth}
    \centering 
    \vspace{-25pt}
 \includegraphics[width=0.34\textwidth]{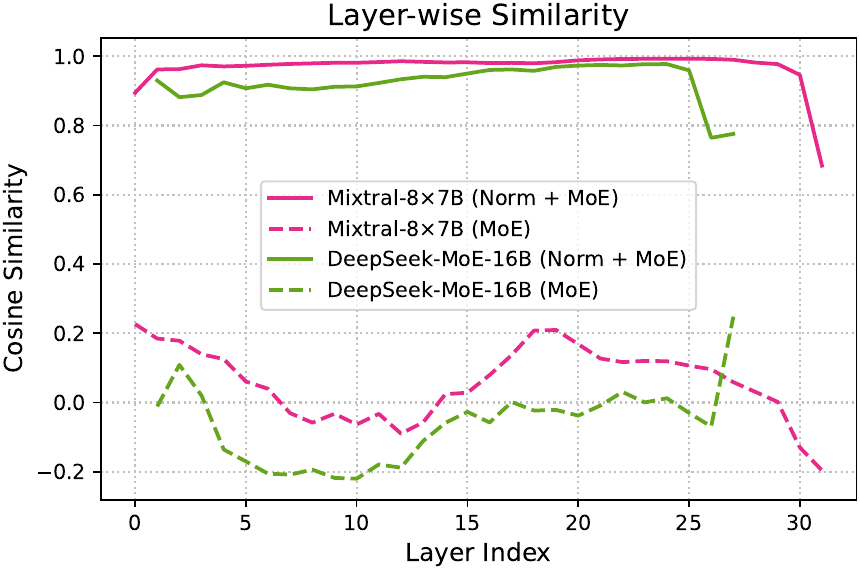} 
    \vspace{-10pt}
    \caption{\textbf{Layer-Wise Similarity.} We consider two scenarios, i.e., for ``MoE'' and ``Norm + MoE''.
    }
    \vspace{-18pt}
    \label{fig:layer_similarity}
\end{wrapfigure}

\begin{figure}[ht]
    \centering
    \includegraphics[width=0.94\linewidth]
    {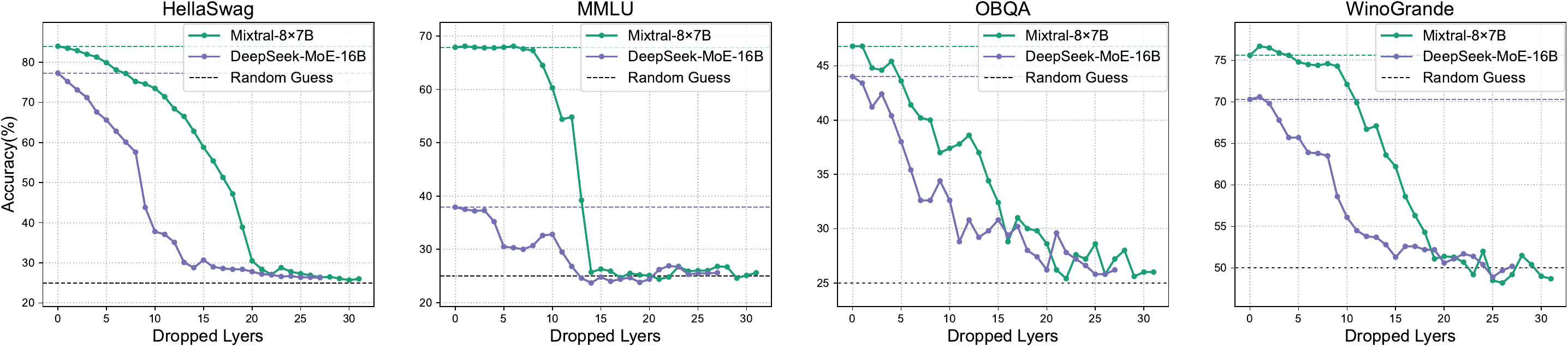} 
    \caption{\textbf{Evaluation of \layerdrop.} We show results on \mixtral~and \deepseek~(solid lines), along with the baseline and random guess performances (dotted lines). }
    \label{fig:layer_drop}
\end{figure}

To verify the feasibility of \layerdrop, we visualize feature similarity across different modules in Figure \ref{fig:layer_similarity}. This visualization shows a high level of similarity for features across the the MoE normalization module (Norm) and the MoE layer. In contrast, the low similarity for features across the MoE layer indicates the infeasibility of removing only MoE layers.
Results from Figure \ref{fig:layer_drop} show that \layerdrop~preserves performance within a wide range of compression ratio, e.g. $1\%$ performance drop on MMLU when dropping 8 layers for \mixtral, revealing significant redundancy in the MoE layers.

\begin{figure}[ht]
    \vspace{-5pt}
    \centering
\includegraphics[width=0.94\linewidth]{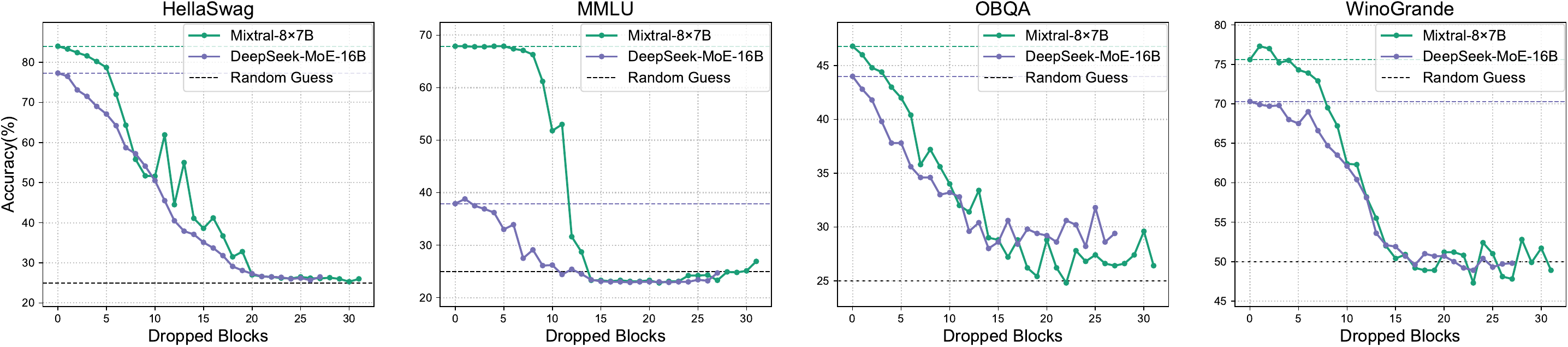} 
    \vspace{-5pt}
    \caption{\textbf{Evaluation of \blockdrop.} We show results on \mixtral~and \deepseek~(solid lines), along with the baseline and random guess performances (dotted lines). }
    \vspace{-10pt}
    \label{fig:block_drop}
\end{figure}

\begin{figure}[ht]
    \centering
    \hspace{-8pt}
    \begin{minipage}{.47\textwidth}
    \includegraphics[width=\textwidth]{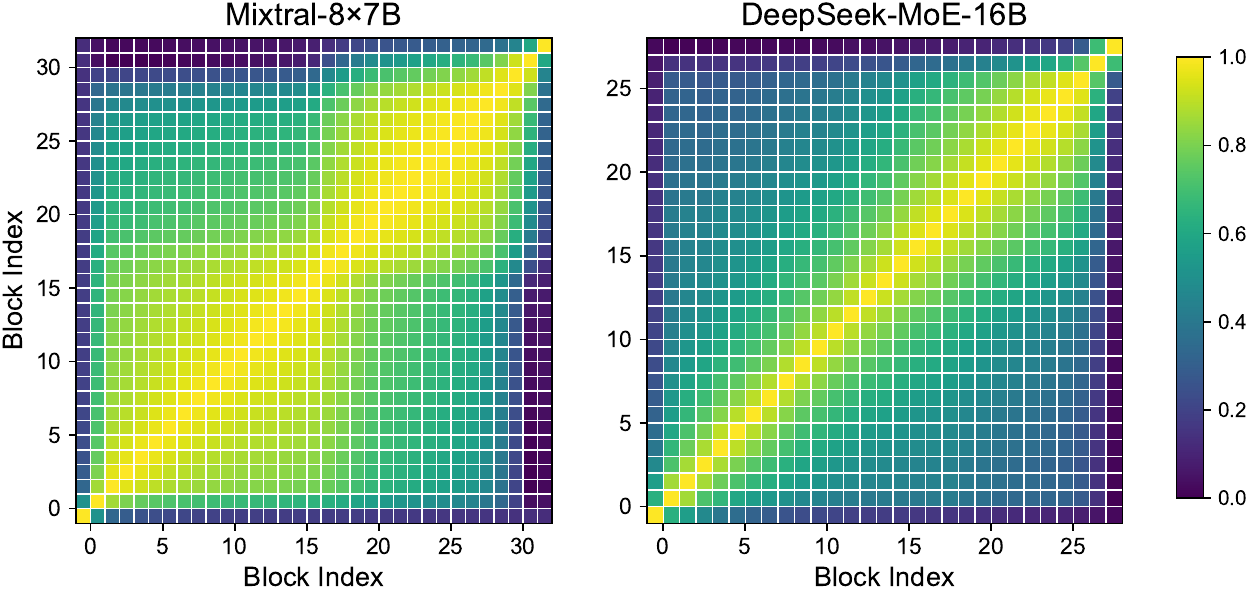}
        \vspace{-12pt}
        \caption{\textbf{Normalized Block-Wise Similarity.} We measure the cosine similarity among hidden features between blocks.}
\label{fig:block_similarity}
    \end{minipage}\hspace{10pt}
    \begin{minipage}{.47\textwidth}
        \includegraphics[width=\textwidth]{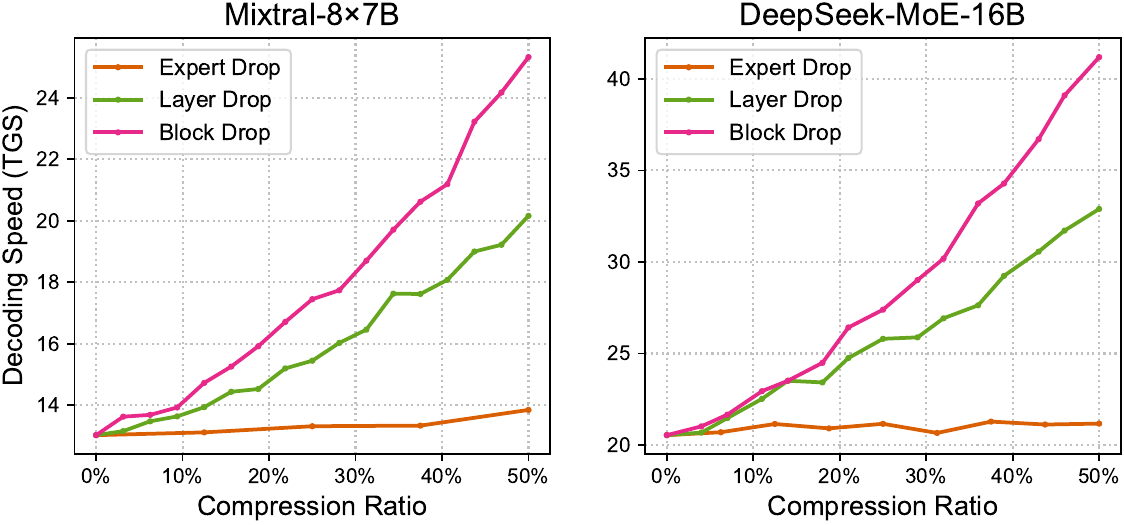} 
        \vspace{-10pt}
        \caption{\textbf{Speedup Scaling Curves of \experttrimming~Methods.} where we measure the averaged decoding speed during generation.} 
        \label{fig:speedup}
    \end{minipage}
    \vspace{-15pt}
\end{figure}


\paragraph{\blockdrop: Further Optimizing Efficiency by Pruning Entire Transformer Blocks}


\begin{wraptable}{HTBPr}{0.42\textwidth}
    \centering
    \caption{\textbf{Comparison of \layerdrop~and \blockdrop~on dense and MoE models.} ``-L$n/m$'', ``-B$n/m$'' represents dropping $n$ out of $m$ corresponding modules with \layerdrop~and \blockdrop, respectively.
    }
    \vspace{-5pt}
    \resizebox{0.41\columnwidth}{!}{
    \setlength{\tabcolsep}{2pt}
    \begin{tabular}{lccccl}
    \toprule
    \multicolumn{6}{c}{\bf Mistral-7B (Dense)} \\
     \midrule
     \bf Method~ & ~ARC-C~ & HellaSwag & ~MMLU~ & ~OBQA~ & ~~~\underline{Average} \\
    \midrule
    Baseline~ & 61.5 & 83.7 & 62.5 & 43.8 & ~
    \underline{62.9} \\
     \midrule
     + L4/32 & 53.2 & 77.7 & 61.7 & 40.0 & 
     ~~\underline{58.2} \small (-4.7) \\
     + L8/32 & 36.7 & 33.6 & 53.3 & 30.6 & 
     ~~\underline{38.6} \small (-24.3) \\
     \hdashline
     + B4/32 & 53.1 & 77.5 & 61.6 & 40.0 & 
     ~~\underline{58.1} \small (-4.8) \\
     + B8/32 & 40.0 & 63.9 & 60.0 & 30.6 & 
     ~~\underline{48.6} \small (-14.3) \\
     \bottomrule
     \toprule
    \multicolumn{6}{c}{\bf \mixtral~(MoE)} \\
     \midrule
     \bf Method~ & ~ARC-C~ & HellaSwag & ~MMLU~ & ~OBQA~ & ~~~\underline{Average} \\
    \midrule
    Baseline~ & 59.4 & 84.0 & 67.9 & 46.8 & 
    ~~\underline{64.6}  \\
    \midrule
     + L4/32 & 56.2 & 81.3 & 67.6 & 44.6 & 
     \bf ~~\underline{62.4} \small (-2.2) \\
     + L8/32 & 47.7 & 75.2 & 67.3 & 40.0 & 
     \bf ~~\underline{57.6} \small (-7.0) \\
     \hdashline
     + B4/32 & 53.8 & 80.2 & 67.9 & 43.0 & 
     ~~\underline{61.2} \small (-3.4)  \\
     + B8/32 & 40.8 & 55.8 & 66.3 & 37.2 & 
     ~~\underline{50.0} \small (-14.6) \\
    \bottomrule
    \end{tabular}}
\vspace{-10pt}
\label{tab:llama3}
\end{wraptable}

While \layerdrop~maintains the performance of the original models, it still preserves the computation-costly attention layers. To address this, \blockdrop~extends \layerdrop~by removing whole transformer blocks, including both MoE and attention layers, further reducing computational and memory costs.
Figure \ref{fig:block_similarity} visualizes block-wise similarity, where both \mixtral~and \deepseek~demonstrate high similarity between specific blocks. Based on this observation, we conduct the empirical study by varying the number of dropped blocks.

Surprisingly, as shown in Figure \ref{fig:block_drop}, the \mixtral~maintains over $90\%$ of the original performance even after removing 5 blocks (over 7 billion parameters). Similar observations are also found in \deepseek, where 4 blocks can be removed when maintaining $90\%$ performance. Since \blockdrop~removes computationally expensive attention layers, it outperforms \layerdrop~by a large margin in terms of both memory and inference cost, as illustrated in Figure \ref{fig:speedup}. 

On the other hand, \blockdrop~prunes attention layers along with their corresponding KV-Cache \cite{pope2022efficiently}. For instance, an input sequence with a batch size of 128 and a sequence length of 2048 results in 32GB of KV-Cache, which can be reduced by 5GB using \blockdrop. Overall, by targeting higher-level structures, \layerdrop~and \blockdrop~achieve substantial efficiency improvements while maintaining acceptable performance levels.


\paragraph{MoE Layers are More Redundant than Dense Counterparts}
Since \layerdrop~and \blockdrop~can also be applied to dense models, we take Mistral-7B, the corresponding dense model of \mixtral~for comparison. Both models have the same depth and differ only in the FFN implementation, so we remove the same number of layers or blocks from each. When dropping an equal number of blocks, both MoE and dense models exhibit performance degradation. 
However, the MoE model suffers less performance drop under the same compression setting. For example, when dropping 8 MoE layers, the Mistral-7B receives a performance drop of 24.3, while \mixtral~ only receives a drop of 7.0. This interesting finding highlights the higher redundancy in MoE layers, and further validates the effectiveness of applying \layerdrop~and \blockdrop~to MoE models.









\section{Visualization Examples of \layerdrop~and \blockdrop}
\label{app:data}

In this section, we visualize the layer-wise similarity and the corresponding dropping order of MoE layers and blocks to investigate the varying levels of redundancy across different depths.

\begin{figure}[htbp]
    \centering
    \begin{subfigure}{0.45\textwidth}
        \includegraphics[width=\textwidth]{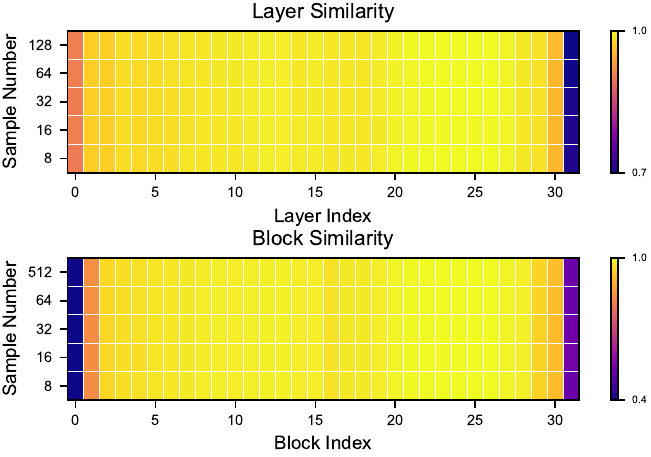} 
        \subcaption{Similarities under different number of samples.}
        \label{fig:ablation_sample_num}
    \end{subfigure}\hspace{7pt}
    \begin{subfigure}{0.45\textwidth}
        \includegraphics[width=\textwidth]{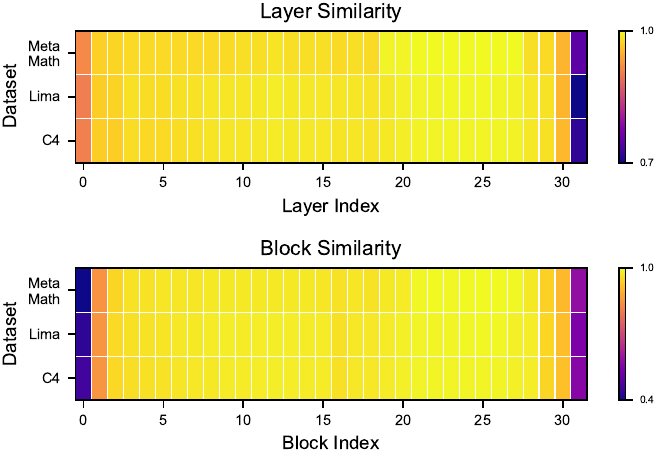} 
        \subcaption{Similarities under different datasets.}
        \label{fig:ablation_sample_type}
    \end{subfigure}
    \vspace{-5pt}
    \caption{\textbf{Influence of Data Choices on Feature Similarity.} We measure the similarity among layers and blocks on \mixtral. \textbf{(a)} The similarity calculated using different number of samples from C4 \cite{2019t5}. \textbf{(b)} The normalized similarity calculated using $1,024$ samples from different datasets, i.e., C4, Lima \cite{zhou2023lima} and MetaMathQA \cite{yu2023metamath}. 
    }
        \vspace{-10pt}
\label{fig:ablation_sample}
\end{figure}

Since our similarity-based metrics depend on the hidden states of each block, the choice of data may influence feature similarity across layers. To investigate this, we conducted ablation studies on \mixtral, examining both the number of samples and the types of datasets used for feature extraction. This analysis helps us understand how data selection affects decisions regarding the dropping of layers or blocks. The results are presented in Figure \ref{fig:ablation_sample}.

\paragraph{Robustness to Calibration Datasets} 
In Figure \ref{fig:ablation_sample_num}, we note that the feature similarity remains relatively stable across different layers as the sample size increases, indicating that \layerdrop~and \blockdrop~maintain consistency regardless of the sample quantity. This confirms that using $128$ samples suffices for computing similarity, which is adopted for all our experiments. Similarly, Figure \ref{fig:ablation_sample_type} shows that varying the datasets, from pretraining with C4 to instruction tuning with Lima and MetaMathQA, does not significantly alter the feature similarity. This demonstrates the resilience of \layerdrop~and \blockdrop~to variations in data distribution.


\begin{figure}[h]
    \centering
    \begin{subfigure}{0.45\textwidth}
        \includegraphics[width=\textwidth]{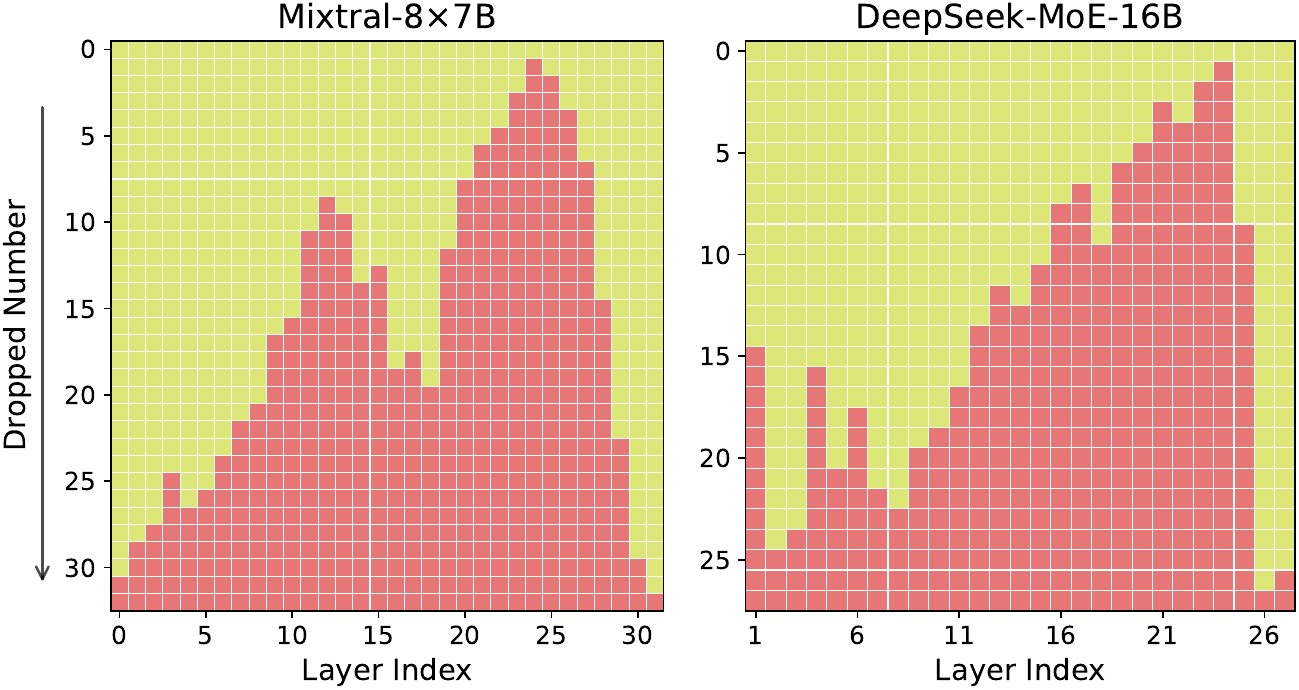} 
        \subcaption{\layerdrop}
    \end{subfigure}
    \hspace{10pt}
    \begin{subfigure}{0.45\textwidth}
        \includegraphics[width=\textwidth]{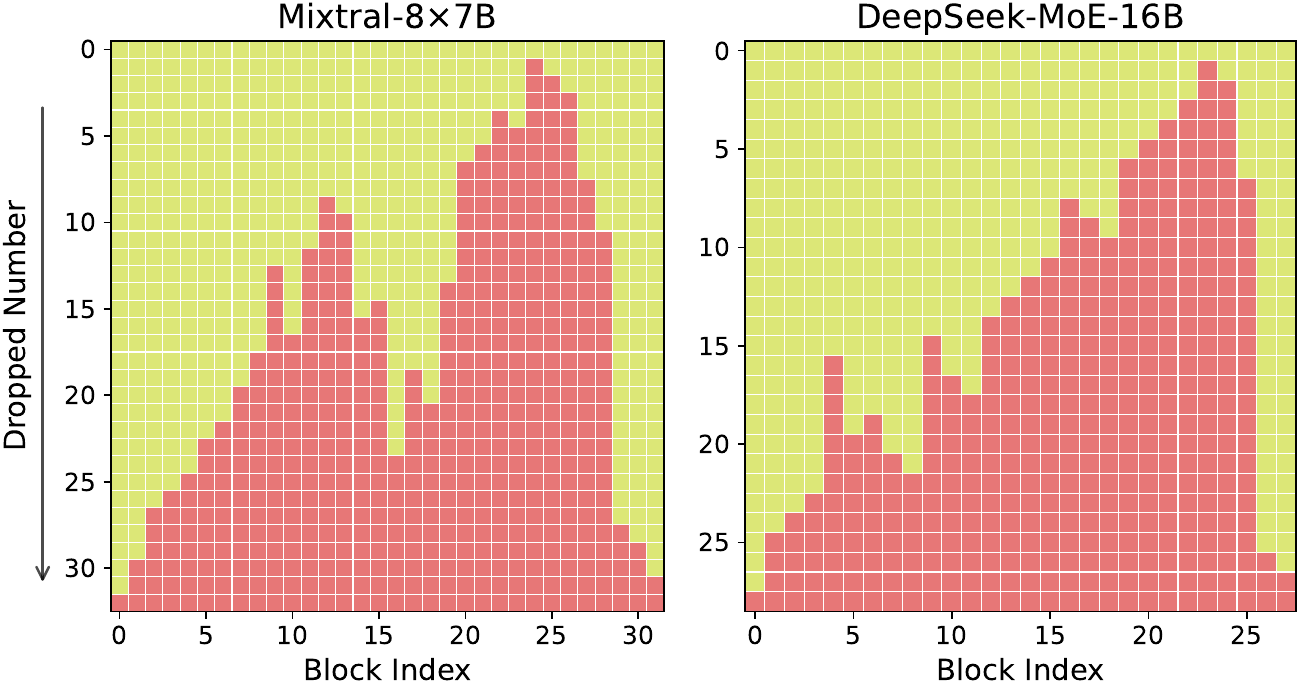} 
        \subcaption{\blockdrop}
    \end{subfigure}
    \vspace{-8pt}
    \caption{
    \textbf{Dropping Patterns for \layerdrop~and \blockdrop.} We visualize of the remaining layers and blocks under different dropped numbers, where yellow areas represent the retained portions and red areas indicate the dropped layers/blocks. }
    \label{fig:drop_order}
    \vspace{-12pt}
\end{figure}

\paragraph{Redundant Deeper Layers} Figure \ref{fig:drop_order} visualizes the remaining and dropped layers/blocks as the number of dropped modules increases. Both MoE architectures exhibit similar patterns in \layerdrop~and \blockdrop: initially, both models tend to drop the deeper layers, followed by the shallower ones. These findings are consistent with Xu \textit{et al.} \cite{men2024shortgpt}, which suggests that deeper layers tend to be more redundant. 



\begin{table}[htbp]
\renewcommand{\arraystretch}{1.0} 
\setlength{\tabcolsep}{2.3pt} 
\centering
\caption{
\textbf{Experimental Results of the Integration of \experttrimming~and \expertslimming}. 
``-E$n/m$'' denotes dropping $n$ out of $m$ experts per MoE layer on average. 
``-L$n/m$'', ``-B$n/m$'' represents dropping $n$ out of $m$ layers/blocks with \layerdrop~and \blockdrop, respectively. 
The speedup quantifies the relative reduction in processing time achieved by the compressed models compared to the baseline. Both the speedup and FLOPs are evaluated by running a forward pass on an input sequence of length $2,048$.
}
\vspace{-5pt}
\resizebox{0.95\linewidth}{!}{
\begin{tabular}{lcccccccccccc}
    \toprule
    \multicolumn{13}{c}{\bf \mixtral} \\ 
    \midrule
    \bf Method & SpeedUp & FLOPs & Memory & ARC-C & BoolQ & HellaSwag & MMLU & OBQA & PIQA & RTE & WinoGrande & \underline{Avg.} \\
    \midrule
    Baseline & -- & 54.4T & 87.7GB
    & 59.4 & 84.2 & 84.0 & 67.9 & 46.8 & 83.8 & 70.4 & 75.6 & \underline{71.5} \\
    \hdashline
    w/AWQ & $5.08 \times$ & 54.4T & 24.4GB
    & 58.4 & 84.2 & 83.3 & 66.6 & 45.8 & 83.0 & 69.0 & 76.3 & \underline{70.8} \\
    \midrule
    + E2/8 & $1.06 \times$ & 54.4T & 66.7GB
    & \bf 53.2 & 77.7 & \bf 80.5 & 52.2 & \bf 46.2 & \bf 81.7 & 55.6 & 76.8 & \underline{65.5} \\
    \hdashline
    w/AWQ & $5.28 \times$ & 54.4T & \bf 20.1GB
    & \bf 50.7 & 79.1 & \bf 78.9 & 52.4 & \bf 44.2 & \bf 81.2 & 55.6 & 75.9 & \underline{64.8} \\
    \midrule
    + L8/32 & $1.19\times$ & \bf 42.9T & 66.6GB
    & 47.7 & 85.3 & 75.2 & \bf 67.3 & 40.0 & 75.8 & \bf 69.7 & \bf 74.6 & \underline{67.0} \\
    \hdashline
    w/AWQ & $\mathbf{6.05\times}$ & \bf 42.9T & \bf 20.0GB
    & 46.2 & 84.2 & 74.2 & \bf 66.2 & 39.0 & 75.5 & \bf 69.3 & \bf 74.2 & \underline{66.1} \\
    \midrule
    + B5/32 & $1.17\times$ & \bf 46.0T & 74.1GB
    & 51.3 & \bf 85.3 & 78.7 & \bf 67.9 & 42.0 & 79.3 & \bf 69.7 & \bf 74.3 & \bf \underline{68.6} \\
    \hdashline
    w/AWQ & $\mathbf{5.94\times}$ & \bf 46.0T & 21.9GB
    & 50.6 & \bf 85.1 & 77.5 & \bf 66.9 & 41.4 & 76.1 & \bf 71.8 & \bf 74.5 & \bf \underline{68.0} \\
    \bottomrule
    \toprule
    \multicolumn{13}{c}{\bf \deepseek} \\ 
    \midrule
    \bf Method & SpeedUp & FLOPs & Memory & ARC-C & BoolQ & HellaSwag & MMLU & OBQA & PIQA & RTE & WinoGrande & \underline{Avg.} \\
    \midrule
    Baseline & -- & 11.7T & 30.8GB
    & 48.1 & 72.4 & 77.3 & 37.9 & 44.0 & 80.4 & 63.9 & 70.3 & \underline{61.8} \\
    \hdashline
    w/AWQ & $3.16\times$ & 11.7T & ~~9.8GB & 46.8 & 71.2 & 76.6 & 36.4 & 43.6 & 80.1 & 62.1 & 70.1 & \underline{60.9} \\
    \midrule
    + E16/64 & $1.06\times$ & 11.7T & 23.9GB
    & \bf 45.0 & 67.1 & \bf 75.6 & 31.8 & \bf 42.2 & \bf 80.2 & \bf 59.9 & \bf 70.0 & \bf \underline{59.0} \\
    \hdashline
    w/AWQ & $3.34\times$ & 11.7T & 
    \bf 
    ~~7.7GB
    & \bf 44.0 & 66.0 & \bf 74.5 & 27.9 & \bf 42.6 & \bf 78.5 & \bf 56.3 & \bf 67.3 & \bf \underline{57.1} \\
    \midrule
    + L4/28 & $1.14\times$ & 
    \bf 
    10.6T & 26.6GB
    & 39.5 & \bf 70.2 & 67.6 & 35.2 & 40.4 & 75.8 & 48.4 & 65.7 & \underline{55.3} \\
    \hdashline
    w/AWQ & $\mathbf{3.60\times}$ & 
    \bf 
    10.6T & \bf ~~8.5GB
    & 42.1 & \bf 72.0 & 69.2 & 33.7 & 39.8 & 75.1 & 47.7 & 66.5 & \underline{55.8} \\
    \midrule
    + B4/28 & $1.16\times$ & 
    \bf 
    10.1T & 26.4GB
    & 40.3 & \bf 71.3 & 69.0 & \bf 36.2 & 37.8 & 75.8 & 51.6 & 68.0 & \underline{56.3} \\ 
    \hdashline
    w/AWQ & $\mathbf{3.67\times}$ & 
    \bf 
    10.1T & \bf ~~8.4GB
    & 40.1 & \bf 70.2 & 68.6 & \bf 36.1 & 38.4 & 76.2 & 51.6 & 66.4 & \underline{56.0} \\
    \bottomrule
    \end{tabular}
}
\label{tab:integration}
\vspace{-8pt}
\end{table}

\section{Integration of \experttrimming~and \expertslimming}
\label{sec:integration}

Beyond \experttrimming, another avenue for MoE compression is \expertslimming, which targets individual expert compression. Techniques such as quantization and network pruning are among the most commonly employed methods. We provide a detailed comparison of network pruning and quantization in Appendix \ref{app:slimming}, where the quantization outperforms both in performance and efficiency.

Since \experttrimming~and \expertslimming~focus on different aspects of compression, we further explore their potential integration. 
Given the superior average performance and practical efficiency of quantization, we use it for \expertslimming. For \experttrimming, we include all three methods to provide a complete comparison. The discussion on the order of applying these two compression techniques is provided in the Appendix \ref{app:order}.

\paragraph{Quantization Preserves the Performance of \experttrimming}
As shown in Table \ref{tab:integration}, the integration of \expertslimming~and \experttrimming~significantly enhances overall efficiency. Quantization can be seamlessly combined with three different levels of dropping, achieving comparable performance.
For instance, after quantization, the average performance of \layerdrop~and \blockdrop~is nearly the same, maintaining more than $90\%$ of the performance of the original models.

\paragraph{The Integration Significantly Enhances Efficiency} 
In Table \ref{tab:integration}, the integration of \experttrimming~and quantization promotes efficiency by a large margin. Different \experttrimming~strategies showcase different advantages. Specifically, \expertdrop~contributes to reducing memory usage, but its speedup is marginal. \layerdrop~and \blockdrop~excel in speedup as illustrated in Figure \ref{fig:speedup}, with \blockdrop~demonstrating both higher performance and greater speedup. Taking into account all settings, the combination of \blockdrop~and quantization offers the best efficiency with comparable performance: a $6.05 \times$ speedup with only $20.0$GB memory usage, while maintaining over $92\%$ of the performance on \mixtral, making it available to be deployed on a NVIDIA RTX 3090 GPU.








\section{Post-Finetuning Recovers the Performance}

While the discussed compression techniques maintains most of the performance of the original models, we further conduct post-finetuning to recover the degraded performance. 
Specifically, for comparison, we full-finetune DeepSeek-MoE-16B and corresponding compressed models on the Alpaca-GPT4 dataset \citep{peng2023instruction} for 3 epochs using a learning rate of 8e-6 with 0.03 warmup ratio and cosine scheduling, where the global batch size is set to 32. As shown in Figure \ref{tab:post-finetuning}, the post-finetuning process significantly reduces the performance gap between the compressed models and the original models, e.g. narrowing it from 5.5\% to 0.6\% for the model following \blockdrop.

\begin{table}[htbp]
\renewcommand{\arraystretch}{1.0} 
\setlength{\tabcolsep}{2.3pt} 
\centering
\caption{
\textbf{Performance of the DeepSeek-MoE-16B models finetuned after \experttrimming}. We mark the relative average performance loss of the compressed models compared to baselines in brackets.
}
\resizebox{0.98\linewidth}{!}{
\begin{tabular}{lcccccccccccl}
    \toprule
    \multicolumn{13}{c}{\bf \deepseek} \\ 
    \midrule
    \bf Method & SpeedUp & FLOPs & Memory & ARC-C & BoolQ & HellaSwag & MMLU & OBQA & PIQA & RTE & WinoGrande & \ \ \underline{Average} \\
    \midrule
    Baseline & \multirow{2}{*}{--} & \multirow{2}{*}{11.7T} & \multirow{2}{*}{30.8GB}
    & \bf 48.1 & 72.4 & 77.3 & 37.9 & 44.0 & \bf 80.4 & 63.9 & 70.3 & \underline{61.8} \\
    +SFT & & &
    & 44.6 & \bf 75.3 & \bf 79.0 & \bf 40.3 & \bf 44.6 & 80.3 & \bf 70.4 & \bf 71.7 & \bf \underline{63.3} \\
    \midrule
    + E16/64 & \multirow{2}{*}{$1.06\times$} & \multirow{2}{*}{11.7T} & \multirow{2}{*}{23.9GB}
    & \bf 45.0 & 67.1 & 75.6 & 31.8 & 42.2 & \bf 80.2 & 59.9 & 70.0 & \underline{59.0} (-2.8) \\
    +SFT & & &
    & 44.4 & \bf 74.0 & \bf 78.6 & \bf 38.5 & \bf 45.8 & 79.6 & \bf 65.7 & \bf 70.1 & \bf \underline{62.1} (-1.2) \\
    \midrule
    + L4/28 & \multirow{2}{*}{$1.14\times$} & \multirow{2}{*}{10.6T} & \multirow{2}{*}{26.6GB}
    & 39.5 & 70.2 & 67.6 & 35.2 & 40.4 & 75.8 & 48.4 & 65.7 & \underline{55.3} (-6.5) \\
    +SFT & & &
    & \bf 42.1 & \bf 78.9 & \bf 75.2 & \bf 40.8 & \bf 43.4 & \bf 77.6 & \bf 71.1 & \bf 69.5 & \bf \underline{62.3} (-1.0) \\
    \midrule
    + B4/28 & \multirow{2}{*}{$1.16\times$} & \multirow{2}{*}{10.1T} & \multirow{2}{*}{26.4GB}
    & 40.3 & 71.3 & 69.0 & 36.2 & 37.8 & 75.8 & 51.6 & 68.0 & \underline{56.3} (-5.5) \\
    +SFT & & &
    & \bf 43.2 & \bf 78.2 & \bf 75.0 & \bf 40.4 & \bf 43.8 & \bf 76.8 & \bf 74.0 & \bf 70.2 & \bf \underline{62.7} (-0.6) \\
    \bottomrule
    \end{tabular}
}
\label{tab:post-finetuning}
\end{table}

\section{Conclusion}
In this paper, we conducted a holistic study of MoE compression techniques, facilitating a systematic understanding of the efficiency issue of MoE and identifying the new design space to improve the performance further.
Based on this study, we propose a comprehensive recipe that integrates \expertslimming~and \experttrimming~to further enhance efficiency. Our proposed methods and insights not only address current challenges but also set the stage for future advancements in the field of MoE.

\bibliography{main}
\bibliographystyle{tmlr}

\appendix
\newpage
\appendix

\newpage

\section{Implementation Details}
\label{app:implementation_details}

\subsection{Models and Datasets}
\paragraph{Models} For our experiments, we employed \mixtral~\citep{jiang2024mixtral} and \deepseek~\citep{dai2024deepseekmoe}. 
\mixtral~utilizes 8 experts for MoE layers and activates the top two for each input token. In contrast, \deepseek~employs a dense FFN in the first block and utilizes two shared experts with an additional 64 experts within MoE layers in other blocks. 

\paragraph{Datasets} For compression experiments, we used the C4 dataset \citep{2019t5}, with 128 samples and an input sequence length of 2,048, following the setup in \citep{wanda, lu2024experts, awq, gptq}. To evaluate model performance, we report normalized zero-shot accuracy on the LM-harness benchmark, which includes multiple tasks: ARC-C \citep{clark2018think}, BoolQ \citep{clark2019boolq}, HellaSwag \citep{zellers2019hellaswag}, MMLU \citep{hendrycks2021measuring}, OBQA \citep{mihaylov2018suit}, PIQA \citep{bisk2019piqa}, RTE \citep{wang2019glue}, and WinoGrande \citep{ai2:winogrande}. The evaluation code is based on EleutherAI LM Harness \citep{eval-harness}.

\subsection{Implementation Details of \expertslimming}
Both \expertslimming~methods (i.e., pruning and quantization) require calibration data to estimate input statistics. To control this variable, we use $128$ samples from the C4 dataset \citep{2019t5} as the calibration dataset for pruning. For quantization, we follow the default settings of GPTQ \footnote{https://github.com/AutoGPTQ/AutoGPTQ} and AWQ \footnote{https://github.com/casper-hansen/AutoAWQ}, using $128$ random samples from Alpaca \citep{alpaca} and Pile \citep{gao2020pile}, respectively. We use the default group size $128$ for \mixtral~and $64$ for \deepseek. 

\subsection{Implementation Details of \expertdrop}
\label{app:expert_drop_details}

The \expertdrop~compresses MoE by preserving only important experts $\{\expert_i\}_{i \in \T'}$ while removing others, where $\T'$ is determined by the importance scores $\{\score(\expert_i)\}_{i\in\T}$. Following \cite{muzio2024seermoe}, we measure the importance scores through the averaged routing scores of a batched data $\X$, i.e., $\{\score(\expert_i)\} = \frac{1}{|\X|} \sum_{\x\in \X} \gate_i(\x)$, and consider two dropping strategies for \expertdrop: layer-wise dropping and global dropping.

\textbf{Layer-wise dropping} removes the same number of experts for each layer. Given the total number of experts $n=|\T|$ and the preserved number of experts $n'=|\T'|<n$ in layer $l$, the preserved expert set $\T'^{(l)}$ is obtained by:

\begin{equation}
    \T'^{(l)} = \{\expert_t^{(l)}\}, \quad\text{where}\quad \score(\expert_t^{(l)}) \in \mathrm{TopK}(\{\score(\expert_i^{(l)})\}_{i=1}^{n}, n').
\end{equation}

\textbf{Global dropping} constrains the total number of preserved experts for the entire model. Given the total number of layers $L$ in the model, the preserved expert set $\T'^{(l)}$ for layer $l$ is obtained by:

\begin{equation}
    \T'^{(l)} = \{\expert_t^{(l)}\}, \quad\text{where}\quad \score(\expert_t^{(l)}) \in \mathrm{TopK}\Big(\bigcup_{j=1}^m\{\score(\expert_i^{(j)})\}_{i=1}^{n}, n'L\Big).
\end{equation}

For the integration of \expertslimming~and \experttrimming, we choose the global dropping as the strategy of \expertdrop, which shows competitive performance compared to the layer dropping for \mixtral~under low dropping ratios, as well as consistent better performance for \deepseek~in Figure \ref{fig:expert_drop_full}.

\newpage
\section{\expertslimming} 
\label{app:slimming}

\paragraph{Pruning: Comparable Performance with Deployment Challenges.}
In Table \ref{tab:pruning}, we evaluate representative pruning algorithms (i.e., Wanda \citep{wanda}, SparseGPT \citep{sparsegpt}) on \mixtral~and \deepseek. Since \deepseek~utilizes both shared experts and normal experts, we conduct an ablation study on whether to prune shared experts, as discussed in Appendix \ref{sec:shared_experts}.
We find that unstructured pruning preserves more than $95\%$ of performance. However, it is not compatible with existing hardware. Conversely, the hardware-friendly semi-structured pruning (i.e., $4$:$8$ and $2$:$4$ patterns) undergoes a significant performance drop. Nevertheless, according to \cite{lu2024experts}, semi-structured sparsity is ineffective in speeding up MoE models.

\begin{table}[htbp]
\renewcommand{\arraystretch}{0.9} 
\centering
\caption{\textbf{Performance of Pruning on MoE.} We consider two mainstream pruning methods (i.e., Wanda \citep{wanda} and SparseGPT \citep{sparsegpt}) under $50\%$ unstructured sparsity and $2$:$4$ semi-structured sparsity.}
\vspace{3pt}
\resizebox{0.95\linewidth}{!}{
\begin{tabular}{lcccccccccc}
    \toprule
    \multicolumn{11}{c}{\bf \mixtral} \\ 
    \midrule
    \bf Method & Sparsity & ARC-C & BoolQ & HellaSwag & MMLU & OBQA & PIQA & RTE & WinoGrande & \underline{Avg.} \\
    \midrule
    Baseline & $0\%$
    & 59.4 & 84.2 & 84.0 & 67.9 & 46.8 & 83.8 & 70.4 & 75.6 & \underline{71.5} \\
    \midrule
    Wanda & 
    & 56.1 & 85.8 & 81.7 & 64.3 & 46.4 & 82.2 & 65.0 & 76.0 & \underline{69.7} \\
    SparseGPT & \multirow{-2}{*}{$50\%$}
    & 56.4 & 85.7 & 81.5 & 64.6 & 45.0 & 82.4 & 66.8 & 75.8 & \underline{69.8} \\
    \hdashline
    Wanda & 
    & 51.4 & 79.4 & 77.8 & 60.3 & 44.0 & 80.7 & 65.3 & 74.1 & \underline{66.6} \\
    SparseGPT & \multirow{-2}{*}{$2$:$4$}
    & 49.2 & 81.0 & 77.6 & 59.2 & 44.0 & 80.6 & 63.9 & 74.8 &  \underline{66.3} \\
    \bottomrule
    \toprule
    \multicolumn{11}{c}{\bf \deepseek} \\ 
    \midrule
    \bf Method & Sparsity & ARC-C & BoolQ & HellaSwag & MMLU & OBQA & PIQA & RTE & WinoGrande & \underline{Avg.} \\
    \midrule
    Baseline & $0\%$
    & 48.1 & 72.4 & 77.3 & 37.9 & 44.0 & 80.4 & 63.9 & 70.3 & \underline{61.8} \\
    \midrule
    Wanda &
    & 43.6 & 74.3 & 72.6 & 31.1 & 43.0 & 79.5 & 58.1 & 69.4 & \underline{59.0} \\
    SparseGPT & \multirow{-2}{*}{$50\%$}
    & 43.9 & 73.5 & 74.0 & 33.8 & 41.4 & 79.0 & 61.0 & 68.3 & \underline{59.4} \\
    \hdashline
    Wanda &
    & 38.2 & 66.1 & 67.5 & 27.6 & 39.4 & 77.0 & 53.8 & 66.7 & \underline{54.5} \\
    SparseGPT & \multirow{-2}{*}{$2$:$4$}
    & 43.1 & 68.9 & 71.6 & 27.6 & 41.6 & 78.3 & 57.4 & 66.6 & \underline{56.9} \\
    \bottomrule
    \end{tabular}
}
\label{tab:pruning}
\end{table}



\paragraph{Quantization: Better Performance and Greater Efficiency}
In Table \ref{tab:quant}, we evaluate the impact of 4-bit quantization on MoE. The quantization offers two major benefits: it maintains the comparable performance of the original models and significantly reduces memory costs. Specifically, the quantized models achieve over $98\%$ of the original performance while using less than $30\%$ of the memory.
When quantized with AWQ \citep{awq}, \mixtral~and \deepseek~achieve impressive speedups of $\times 5.08$ and $\times 3.16$, respectively. This demonstrates that 4-bit quantization is effective for deploying MoE models in resource-constrained environments.

\begin{table}[htbp]
    \renewcommand{\arraystretch}{0.9} 
    \centering
    \caption{\textbf{Performance of Quantization on MoE.} We utilize GPTQ \citep{gptq} and AWQ \citep{awq} as the quantization methods for 4-bit compression. }
    \resizebox{0.95\linewidth}{!}{
    \begin{tabular}{lcccccccccccc}
        \toprule
        \multicolumn{12}{c}{\bf \mixtral} \\ 
        \midrule
        \bf Method & Bits & Memory & ARC-C & BoolQ & HellaSwag & MMLU & OBQA & PIQA & RTE & WinoGrande & \underline{Avg.} \\
        \midrule
        Baseline & $16$ & 87.7GB 
        & 59.4 & 84.2 & 84.0 & 67.9 & 46.8 & 83.8 & 70.4 & 75.6 & \underline{71.5} \\
        \hdashline
        GPTQ &  &  
        & 59.0 & 84.4 & 83.4 & 67.1 & 45.2 & 83.1 & 70.1 & 75.2 & \underline{70.9} \\
        AWQ & \multirow{-2}{*}{$4$} & \multirow{-2}{*}{24.4GB}
        & 58.4 & 84.2 & 83.3 & 66.6 & 45.8 & 83.0 & 69.0 & 76.3 & \underline{70.8} \\
        \bottomrule
        \toprule
        \multicolumn{12}{c}{\bf \deepseek} \\ 
        \midrule
        \bf Method & Bits & Memory & ARC-C & BoolQ & HellaSwag & MMLU & OBQA & PIQA & RTE & WinoGrande & \underline{Avg.} \\
        \midrule
        Baseline & $16$ & 30.8GB 
        & 48.1 & 72.4 & 77.3 & 37.9 & 44.0 & 80.4 & 63.9 & 70.3 & \underline{61.8} \\
        \hdashline
        GPTQ &  &  
        & 46.3 & 71.8 & 76.8 & 36.4 & 43.4 & 80.0 & 63.9 & 70.2 & \underline{61.1} \\
        AWQ & \multirow{-2}{*}{$4$} & \multirow{-2}{*}{~~9.8GB}
        & 46.8 & 71.2 & 76.6 & 36.4 & 43.6 & 80.1 & 62.1 & 70.1 & \underline{60.9} \\
        \bottomrule
        \end{tabular}
    }
    \vspace{-10pt}
    \label{tab:quant}
\end{table}

\newpage
\section{Analysis on the Dropping Patterns of \expertdrop}

\begin{wrapfigure}{R}{0.50\textwidth}
    \vspace{-12pt}
    \includegraphics[width=0.48\textwidth]{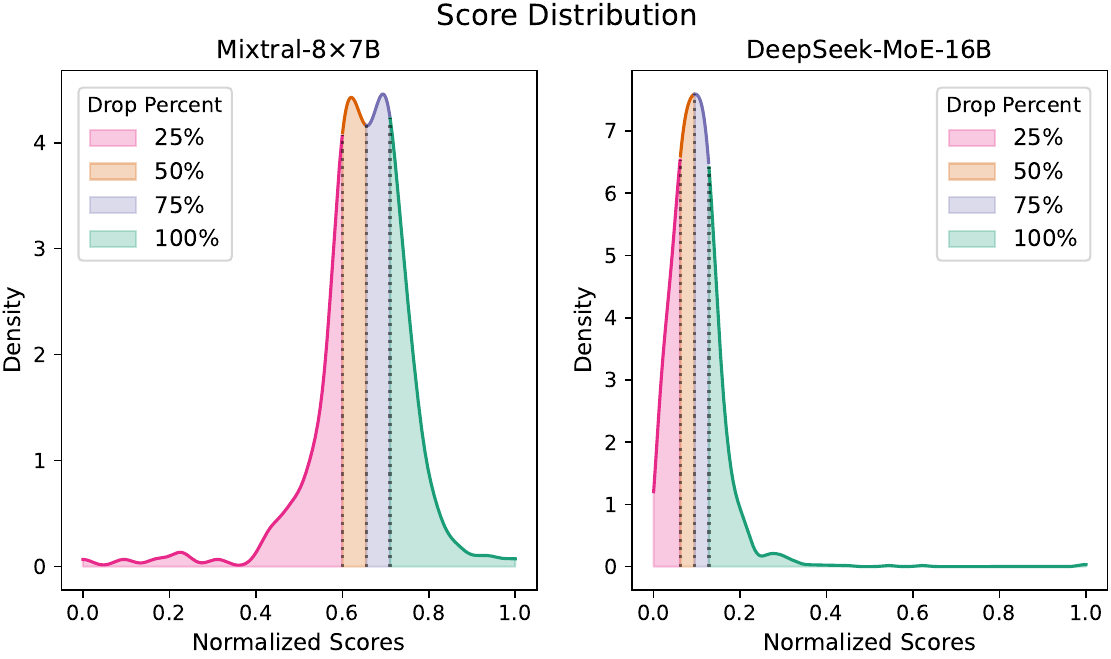}
    \centering
    \vspace{-8pt}
    \caption{\textbf{Distribution of Normalized Importance Scores $\score$ for \expertdrop.} We highlight the density of scores under different drop ratios with different colors.}
        \vspace{-21pt}

    \label{fig:expert_scores_distribution}
\end{wrapfigure}
\paragraph{Score Distribution Directs \expertdrop}
The distribution of importance scores is informative to determine the proportion of dropped experts. In Figure \ref{fig:expert_scores_distribution}, we visualize the score distribution of \expertdrop~for \mixtral~and \deepseek. \deepseek, which allocates more experts, shows a left-skewed distribution where most experts have low scores. In contrast, \mixtral~demonstrates a right-skewed distribution, with only a few experts being deemed unimportant. This distribution difference results in different resistance capabilities against \expertdrop, where \deepseek~can drop more experts than \mixtral~while maintaining competitive performance, as demonstrated in Table \ref{tab:integration} and Figure \ref{fig:expert_drop_full}.  


\begin{wrapfigure}{R}{0.50\textwidth}
    \centering
    \vspace{-12pt}
    \includegraphics[width=0.42\textwidth]{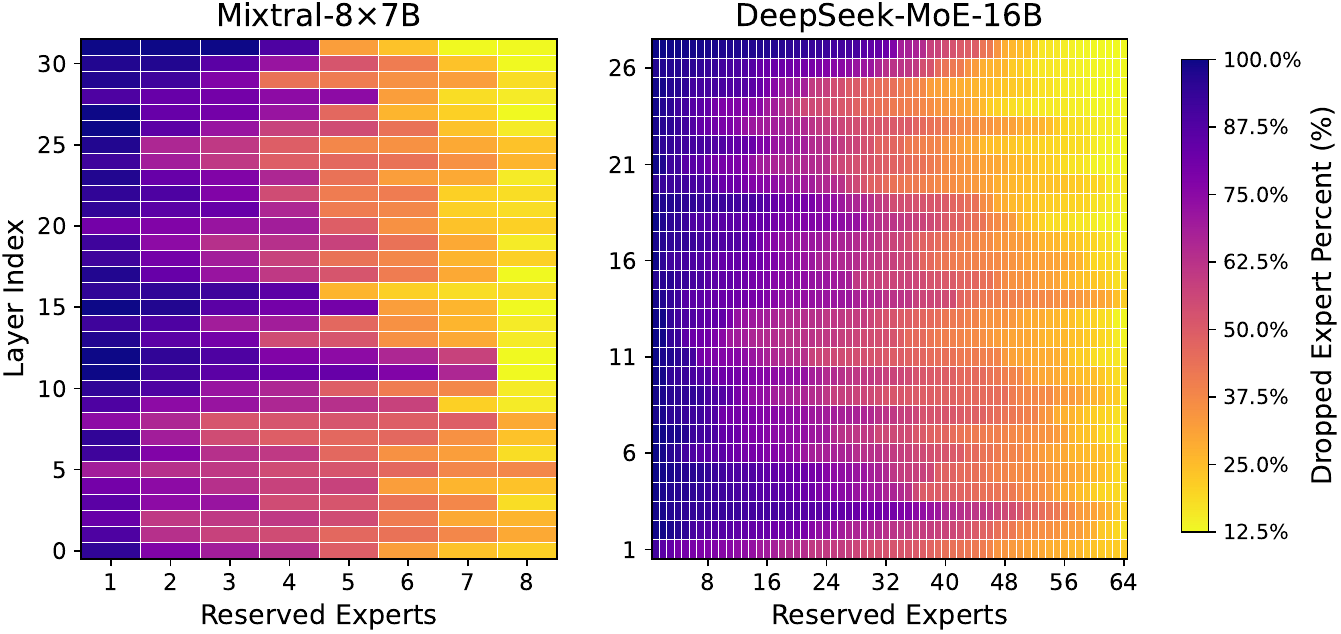} 
    \vspace{-5pt}
    \caption{\textbf{Distribution of Dropped Experts for \expertdrop.}  We visualize the dropped experts under different drop ratios, where the dropped experts are colored from \textbf{\textcolor{yellow}{yellow}} to \textbf{\textcolor{blue}{blue}} as the drop ratio increases.}
    \vspace{-15pt}
    \label{fig:expert_drop_distribution}
\end{wrapfigure}
\paragraph{Global \expertdrop~Removes Experts Fine-Grainedly}
We employed two different strategies for \expertdrop, namely layer-wise and global. Layer-wise dropping treats each layer equally by dropping the same number of experts, while global dropping results in different proportions of remaining experts across layers. We visualize the distribution of remaining experts after global dropping in Figure \ref{fig:expert_drop_distribution}. We find the global dropping shows a more fine-grained pattern on dropping experts, where the bottom layers are more vulnerable under lower dropping ratios (yellow part).


\clearpage

\section{Ablation Study on Compression Orders}
\label{app:order}
In Section \ref{sec:integration}, we discussed the combination of \experttrimming~and \expertslimming. Here we ablate on the orders of compression when combining these two techniques. Results in Table \ref{tab:order} show that the order of \experttrimming~and \expertslimming~doesn't have a significant influence on the performance, where applying \expertslimming~then \experttrimming~(``S+T'') performs slightly better for \mixtral~(e.g. +0.5, +0.4 and +0.1 for \expertdrop, \layerdrop~and \blockdrop, respectively). To this end, we choose ``S+T'' as the final implementation in our experiments.

\begin{table}[htbp]
\vspace{8pt}
\renewcommand{\arraystretch}{1.1} 
\setlength{\tabcolsep}{3.8pt} 
\centering
\caption{\textbf{Ablation results on different orders of \expertslimming~and \experttrimming.} ``S+T'' denotes first applying \expertslimming~then \experttrimming, and ``T+S'' denotes the reversed order.}
\vspace{2pt}
\resizebox{0.94\linewidth}{!}{
\begin{tabular}{lccccccccc}
    \toprule
    \multicolumn{10}{c}{\textbf{Mixtral-8×7B}} \\
    \midrule
    \bf Method & ARC-C & BoolQ & HellaSwag & MMLU & OBQA & PIQA & RTE & WinoGrande & \underline{Avg.} \\
    \midrule
    Baseline             & 59.4  & 84.2  & 84.0  & 67.9 & 46.8 & 83.8 & 70.4 & 75.6 & \underline{71.5} \\
    \midrule
    + E2/8, AWQ (S+T)    & 50.7  & 79.1  & \bf 78.9  & \bf 52.4 & 44.2 &\bf  81.2 & \bf 55.6 & \bf 75.9 & \bf \underline{64.8} \\
    + E2/8, AWQ (T+S)    & \bf 50.8  & \bf 79.9  & 78.7  & 49.2 & \bf 44.4 & 80.9 & 55.2 & 75.4 & \underline{64.3} \\
    \hdashline
    + L8/32, AWQ (S+T)   & 46.2  & 84.2  & \bf 74.2  & \bf 66.2 & 39.0 & \bf 75.5 & \bf 69.3 & \bf 74.2 & \bf \underline{66.1} \\
    + L8/32, AWQ (T+S)   & \bf 46.8  & \bf 84.4  & 74.0  & 65.3 & \bf 39.8 & 75.0 & 66.8 & 73.2 & \underline{65.7} \\
    \hdashline
    + B5/32, AWQ (S+T)   & \bf 50.6  & \bf 85.1  & \bf 77.5  & \bf 66.9 & 41.4 & 76.1 & \bf 71.8 & \bf 74.5 & \bf \underline{68.0} \\
    + B5/32, AWQ (T+S)   & 50.3  & 84.7  & 77.4  & 65.8 & \bf 42.0 & \bf 78.8 & 70.4 & 74.0 & \underline{67.9} \\
    \bottomrule
    \toprule
    \multicolumn{10}{c}{\textbf{\deepseek}} \\ 
    \midrule
    \bf Method & ARC-C & BoolQ & HellaSwag & MMLU & OBQA & PIQA & RTE & WinoGrande & \underline{Avg.} \\
    \midrule
    Baseline             & 48.1  & 72.4  & 77.3  & 37.9 & 44.0 & 80.4 & 63.9 & 70.3 & \underline{61.8} \\
    \midrule
    + E16/64, AWQ (S+T)  & 44.0  & \bf 66.0  & \bf 74.5  & 27.9 & \bf 42.6 & 78.5 & \bf 56.3 & 67.3 & \bf \underline{57.1} \\
    + E16/64, AWQ (T+S)  & \bf 44.7  & 64.1  & 74.0  & \bf 29.0 & \bf 42.6 & \bf 79.9 & 54.2 & \bf 68.4 & \bf \underline{57.1} \\
    \hdashline
    + L4/28, AWQ (S+T)   & 42.1  & \bf 72.0  & \bf 69.2  & \bf 33.7 & 39.8 & \bf 75.1 & \bf 47.7 & \bf 66.5 & \bf \underline{55.8} \\
    + L4/28, AWQ (T+S)   & \bf 42.4  & 71.7  & 69.1  & 33.4 & \bf 40.1 & 74.8 & 47.6 & 66.2 & \underline{55.7} \\
    \hdashline
    + B4/28, AWQ (S+T)   & 40.1  & \bf 70.2  & 68.6  & \bf 36.1 & 38.4 & \bf 76.2 & \bf 51.6 & 66.4 & \underline{56.0} \\
    + B4/28, AWQ (T+S)   & \bf 41.6  & 69.4  & \bf 69.1  & 35.8 & \bf 38.6 & \bf 76.2 & 50.9 & \bf 67.0 & \bf \underline{56.1} \\
    \bottomrule
\end{tabular}
}
\label{tab:order}
\vspace{-8pt}
\end{table}

\newpage
\section{Ablation Study on Shared Experts in \deepseek}
\label{sec:shared_experts}

Although most MoE models follow Equation \ref{eq:moe} to implement the experts, models like \deepseek~adopt a residual \citep{rajbhandari2022deepspeed} form of experts, which brings a special scenario to discuss. In the residual MoE, an additional set of $m$ shared experts $\left\{\expertshared_1, \expertshared_2, \dots, \expertshared_m\right\}$ is always selected by the router $\gate$ and activated for all inputs. Given an input $\x$, the output can be represented as a degenerated form of Equation \ref{eq:moe}, where the scores of shared experts are fixed to $1$:
\begin{equation}
    \y = \sum\nolimits_{i\in\mathcal{K}} {\gate(\x)_i \cdot \expert_i(\x)} + \sum\nolimits_{j=1}^{m} {\expertshared_j(\x)}.
\label{eq:moe_residual}
\end{equation}
This special form of expert routing may bring a difference in the redundancy distribution of MoE. Here we discuss the influence of shared experts through pruning and present the results in Table \ref{tab:pruning_shared_experts}. We find that pruning without the shared experts will boost the performance at a considerable scale, i.e., $+3.6\%$ and $+1.5\%$ of the averaged accuracy for unstructured pruning with Wanda and SparseGPT, respectively. This finding reveals a different pattern of inner redundancy in which the shared experts are less compressible compared to the others in residual MoE models, which may inform future work.

\begin{table}[htbp]
\centering
\caption{
\textbf{Ablation Study of Pruning Shared Experts on \deepseek.} We consider two scenarios, i.e., pruning both shared experts and normal experts (``w/Pruning Shared Experts'') and pruning normal experts only (``w/o Pruning Shared Experts''). We use two mainstream pruning methods (i.e., Wanda \citep{wanda} and SparseGPT \citep{sparsegpt}) under both unstructured sparsity ($50\%$) and semi-structured sparsity ($2$:$4$). 
}
\resizebox{0.98 \linewidth}{!}{
\begin{tabular}{lcccccccccc}
    \toprule
    \multicolumn{11}{c}{\bf \deepseek} \\ 
    \midrule
    \bf Method & Sparsity & ARC-C & BoolQ & HellaSwag & MMLU & OBQA & PIQA & RTE & WinoGrande & \underline{Avg.} \\
    \midrule
    Baseline & $0\%$
    & 48.1 & 72.4 & 77.3 & 37.9 & 44.0 & 80.4 & 63.9 & 70.3 & \underline{61.8} \\
    \midrule
    \multicolumn{11}{c}{
    w/ Pruning Shared Experts
    } 
    \\ 
    \midrule
    Wanda &
    & 43.6 & 74.3 & 72.6 & 31.1 & 43.0 & 79.5 & 58.1 & 69.4 & \underline{59.0} \\
    SparseGPT & \multirow{-2}{*}{$50\%$}
    & 43.9 & 73.5 & 74.0 & 33.8 & 41.4 & 79.0 & 61.0 & 68.3 & \underline{59.4} \\
    \hdashline
    Wanda &
    & 38.2 & 66.1 & 67.5 & 27.6 & 39.4 & 77.0 & 53.8 & 66.7 & \underline{54.5} \\
    SparseGPT & \multirow{-2}{*}{$2$:$4$}
    & 43.1 & 68.9 & 71.6 & 27.6 & 41.6 & 78.3 & 57.4 & 66.6 & \underline{56.9} \\
    \midrule
    \multicolumn{11}{c}{
    w/o Pruning Shared Experts
    } 
    \\ 
    \midrule
    Wanda &
    & 44.0 & 76.3 & 73.5 & 36.2 & 41.0 & 79.3 & 59.9 & 70.2 & \underline{60.0} \\
    SparseGPT & \multirow{-2}{*}{$50\%$}
    & 45.0 & 75.5 & 74.4 & 36.3 & 41.0 & 79.4 & 64.3 & 69.3 & \underline{60.7} \\
    \hdashline
    Wanda &
    & 40.1 & 75.7 & 69.9 & 33.5 & 40.0 & 77.9 & 58.8 & 68.6 & \underline{58.1} \\
    SparseGPT & \multirow{-2}{*}{$2$:$4$}
    & 40.7 & 75.7 & 69.9 & 33.3 & 39.0 & 77.7 & 61.4 & 69.4 & \underline{58.4} \\
    \bottomrule
    \end{tabular}
}
\label{tab:pruning_shared_experts}
\end{table}

\newpage
\section{Full Experimental Results}
\label{sec:full_results}
\begin{figure}[ht]
    \centering
    \includegraphics[width=\linewidth]{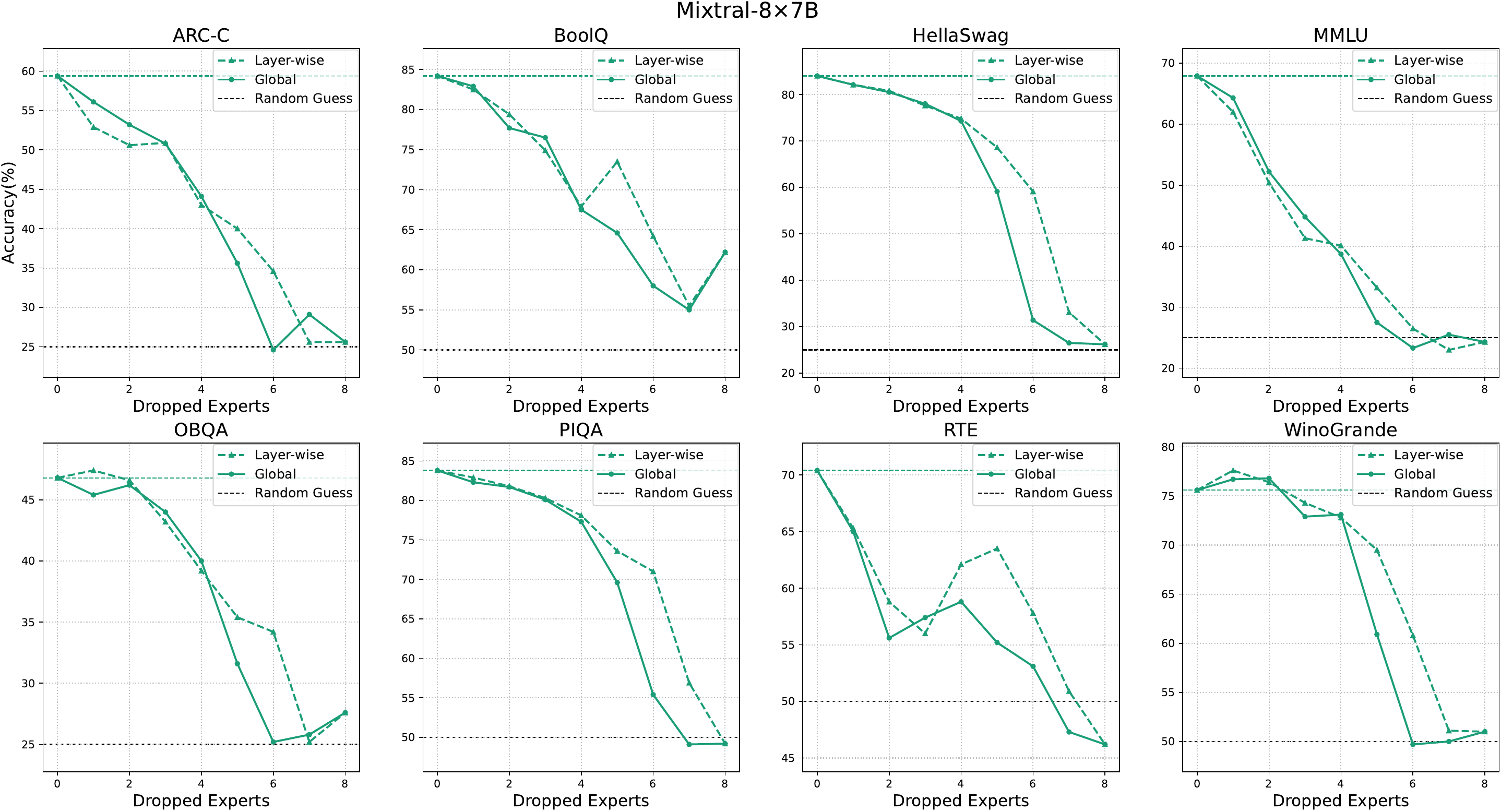}
    \vspace{5pt}
    \includegraphics[width=\linewidth]{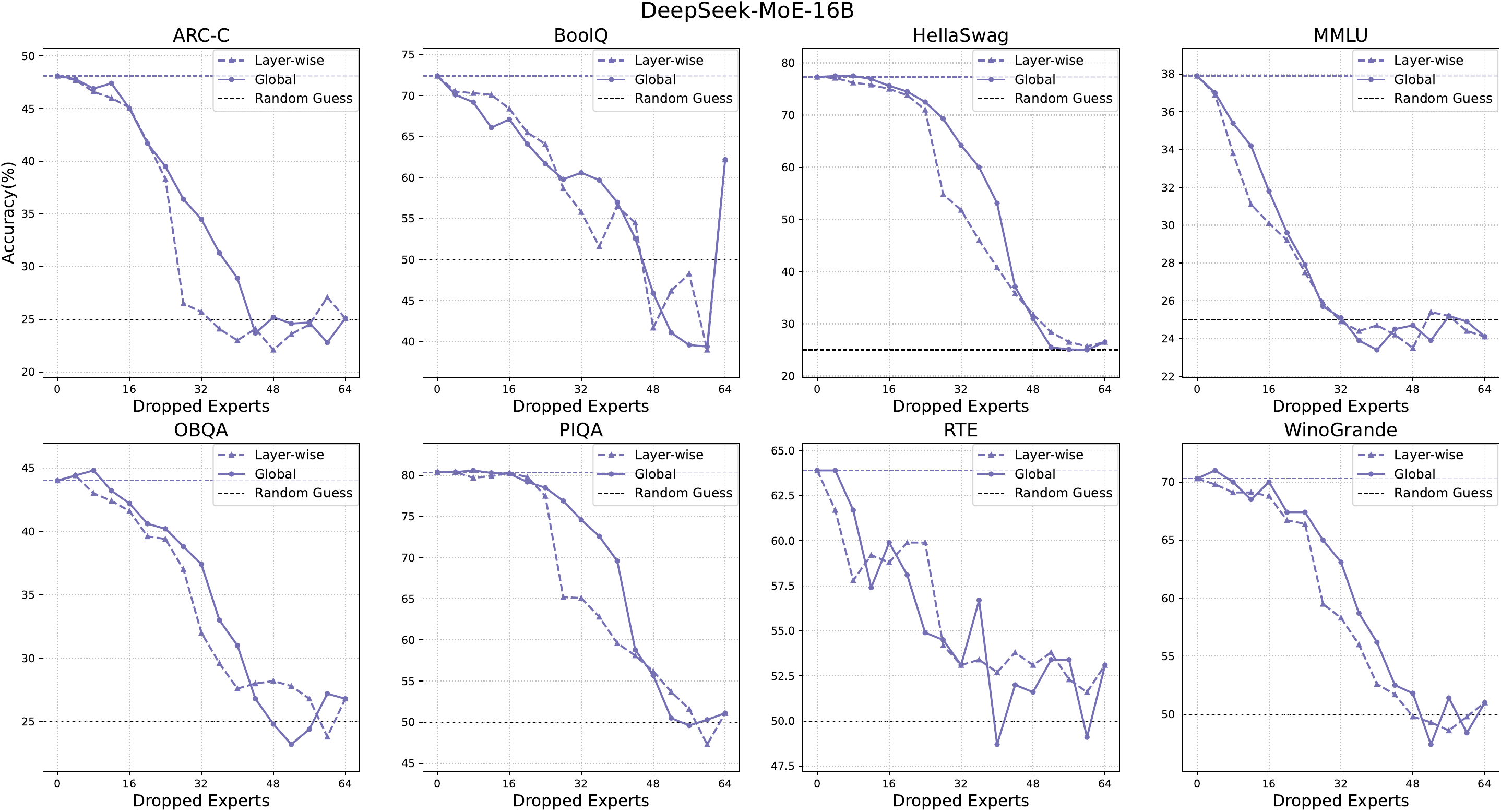} 
    \caption{\textbf{Full Results for \expertdrop.} We consider two strategies: layer-wise (dotted lines) and global (solid lines).}
    \label{fig:expert_drop_full}
\end{figure}

\begin{figure}[ht]
    \centering
    \includegraphics[width=\linewidth]{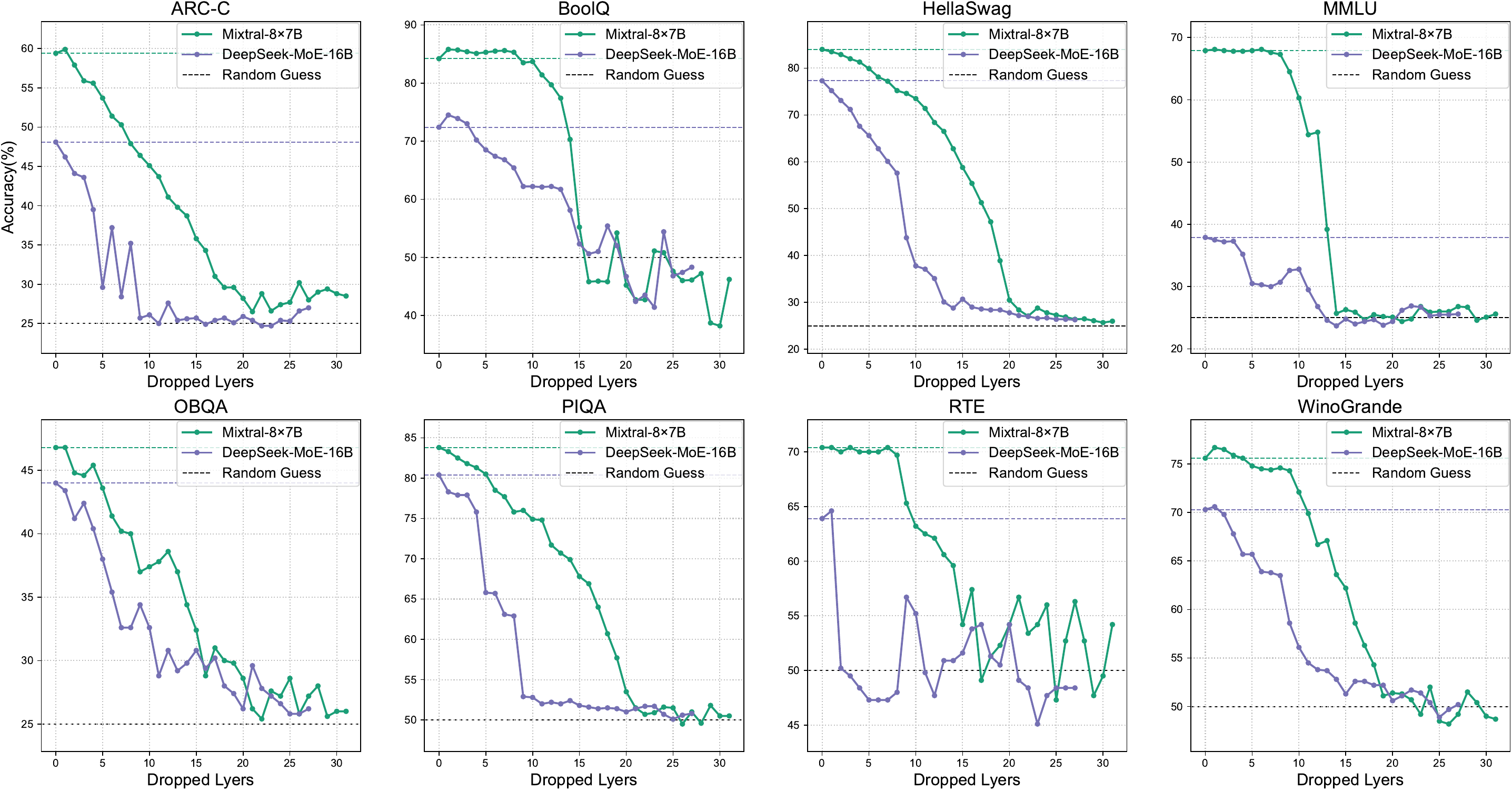} 
    \caption{\textbf{Full Results for \layerdrop.} We show results on \mixtral~and \deepseek~(solid lines), along with the baseline and random guess performances (dotted lines). }
    \label{fig:layer_drop_full}
\end{figure}

\begin{figure}[ht]
    \centering
    \includegraphics[width=\linewidth]{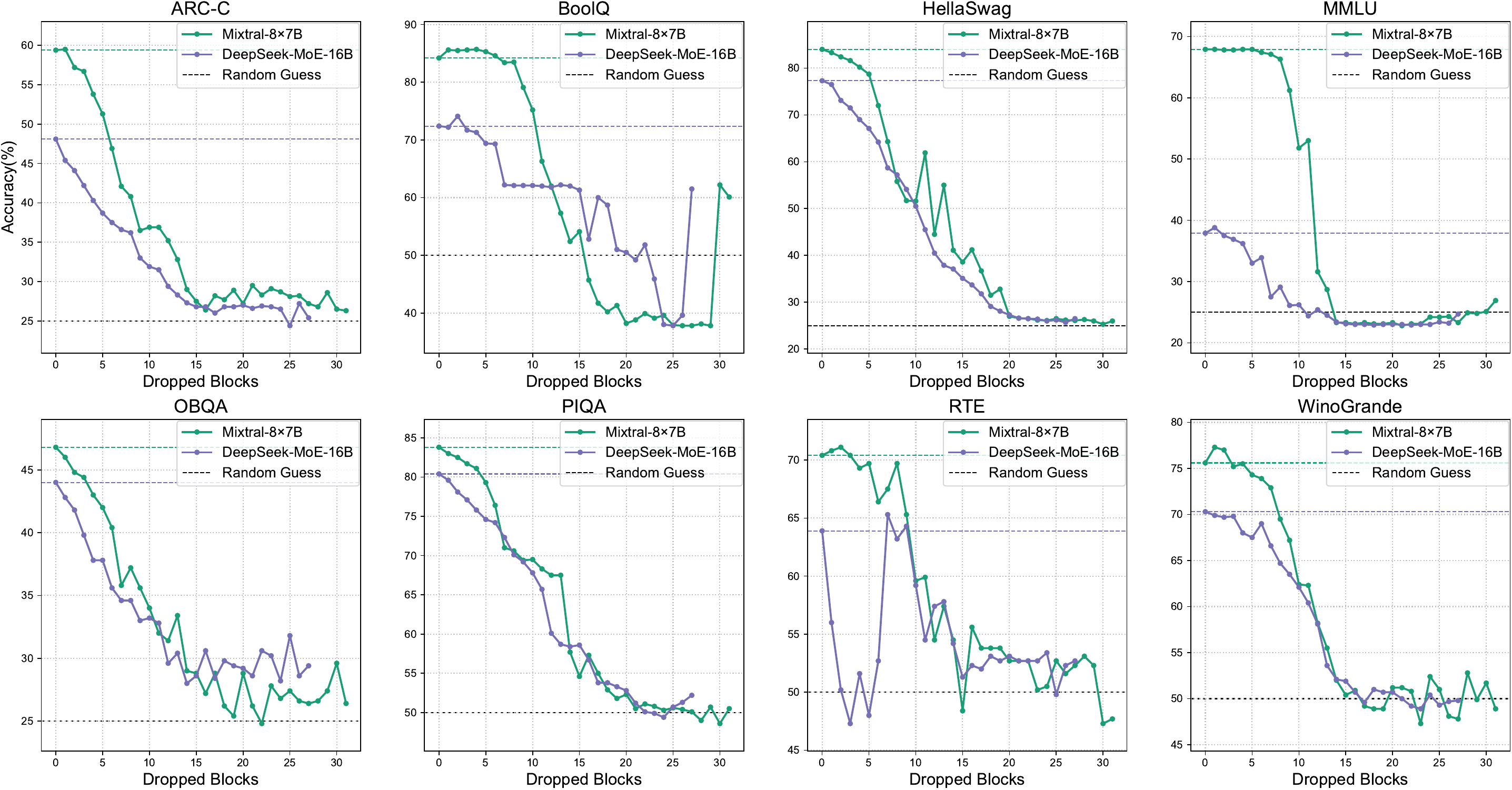} 
    \caption{\textbf{Full Results for \blockdrop.} We show results on \mixtral~and \deepseek~(solid lines), along with the baseline and random guess performances (dotted lines). }
    \label{fig:block_drop_full}
\end{figure}

We provide the full results of \experttrimming, including \expertdrop, \layerdrop~and \blockdrop, in Figure \ref{fig:expert_drop_full}, \ref{fig:layer_drop_full}, and \ref{fig:block_drop_full}, respectively.

\end{document}